\documentclass[pdflatex,sn-basic,iicol]{sn-jnl}
\jyear{2021}%

\theoremstyle{thmstyleone}%
\theoremstyle{thmstyletwo}%

\theoremstyle{thmstylethree}%

\usepackage{bm}
\usepackage{multirow}

\usepackage{amssymb}
\usepackage{pifont}
\newcommand{\cmark}{\ding{51}}%
\newcommand{\xmark}{\ding{55}}%

\begin{document}

\title[ARES-Bench]{A Comprehensive Study on Robustness of Image Classification Models: Benchmarking and Rethinking}

\author[2]{\fnm{Chang} \sur{Liu}}\email{sunrise6513@sjtu.edu.cn}
\equalcont{These authors contributed equally to this work.}

\author[1,5]{\fnm{Yinpeng} \sur{Dong}}\email{dongyinpeng@tsinghua.edu.cn}
\equalcont{These authors contributed equally to this work.}

\author[3,7]{\fnm{Wenzhao} \sur{Xiang}}\email{xiangwenzhao22@mails.ucas.ac.cn}

\author[1]{\fnm{Xiao} \sur{Yang}}\email{yangxiao19@mails.tsinghua.edu.cn}

\author*[1,6]{\fnm{Hang} \sur{Su}}\email{suhangss@tsinghua.edu.cn}

\author[1,5]{\fnm{Jun} \sur{Zhu}}\email{dcszj@tsinghua.edu.cn}

\author[4]{\fnm{Yuefeng} \sur{Chen}}\email{yuefeng.chenyf@alibaba-inc.com}

\author[4]{\fnm{Yuan} \sur{He}}\email{heyuan.hy@alibaba-inc.com}

\author[4]{\fnm{Hui} \sur{Xue}}\email{hui.xueh@alibaba-inc.com}

\author[2]{\fnm{Shibao} \sur{Zheng}}\email{sbzh@sjtu.edu.cn}

\affil*[1]{\orgdiv{Dept. of Comp. Sci. and Tech., Institute for AI, Tsinghua-Bosch Joint ML Center,
THBI Lab, BNRist Center}, \orgname{Tsinghua University}, \orgaddress{\city{Beijing}, \postcode{100084}, \country{China}}}

\affil[2]{\orgdiv{Institute of Image Communication and Networks Engineering
in the Department of Electronic Engineering~(EE)}, \orgname{Shanghai Jiao Tong University}, \orgaddress{\city{Shanghai}, \postcode{200240}, \country{China}}}

\affil[3]{\orgdiv{Key Laboratory of Intelligent Information Processing of Chinese Academy of Sciences (CAS),
Institute of Computing Technology}, \orgname{CAS}, \orgaddress{\city{Beijing}, \postcode{100190}, \country{China}}}

\affil[4]{\orgname{Alibaba Group}, \orgaddress{\city{Hangzhou}, \postcode{310023}, \state{Zhejiang}, \country{China}}}
\affil[5]{\orgname{RealAI}, \orgaddress{\city{Beijing}, \postcode{100085}, \country{China}}}
\affil[6]{\orgname{Zhongguancun Laboratory}, \orgaddress{\city{Beijing}, \postcode{100080}, \country{China}}}
\affil[7]{\orgname{Peng Cheng Laboratory}, \orgaddress{\city{Shenzhen}, \postcode{518000}, \country{China}}}

\abstract{The robustness of deep neural networks is usually lacking under adversarial examples, common corruptions, and distribution shifts, which becomes an important research problem in the development of deep learning. Although new deep learning methods and robustness improvement techniques have been constantly proposed, the robustness evaluations of existing methods are often  inadequate due to their rapid development, diverse noise patterns, and simple evaluation metrics. Without thorough robustness evaluations, it is hard to understand the advances in the field and identify the effective methods. In this paper, we establish a comprehensive robustness benchmark called \textbf{ARES-Bench} on the image classification task. In our benchmark, we evaluate the robustness of 55 typical deep learning models on ImageNet with diverse architectures (e.g., CNNs, Transformers) and learning algorithms (e.g., normal supervised training, pre-training, adversarial training) under numerous adversarial attacks and out-of-distribution (OOD) datasets. Using robustness curves as the major evaluation criteria, we conduct large-scale experiments and draw several important findings, including: 1) there is an inherent trade-off between adversarial and natural robustness for the same model architecture; 2) adversarial training effectively improves adversarial robustness, especially when performed on Transformer architectures; 3) pre-training significantly improves natural robustness based on more training data or self-supervised learning.
Based on ARES-Bench, we further analyze the training tricks in large-scale adversarial training on ImageNet. By designing the training settings accordingly, we achieve the new state-of-the-art adversarial robustness. We have made the benchmarking results and code platform publicly available. 
}


\maketitle

\section{Introduction}\label{sec1}

Deep learning has achieved unprecedented success in numerous computer vision tasks, including image classification \citep{krizhevsky2012imagenet,he2015deep,dosovitskiy2021image}, object detection \citep{ren2015faster,redmon2016you,he2017mask}, etc.
However, these well-performing models are criticized for their lack of robustness to adversarial noises \citep{szegedy2013intriguing,goodfellow2014explaining}, common image corruptions \citep{hendrycks2018benchmarking}, and various types of real-world distribution shifts \citep{barbu2019objectnet,hendrycks2021many,hendrycks2021natural,dong2022viewfool}.
For example, the widely studied \emph{adversarial examples}, generated by applying imperceptible perturbations to natural examples, can lead to erroneous predictions of a target model.
The problems with the robustness of deep learning have become a formidable obstacle to human-level performance.

As deep learning models have been increasingly used in security-sensitive applications (e.g., autonomous driving, medical image processing, etc.), the study of model robustness has become an important research topic in computer vision and machine learning, which spawns a large number of adversarial attack and defense algorithms \citep{Madry2017Towards,Wong2018Provable,dong2018boosting,liao2017Defense,cohen2019certified,zhang2019theoretically,croce2020reliable,pang2020bag}, as well as out-of-distribution (OOD) datasets \citep{hendrycks2018benchmarking,geirhos2018imagenet,barbu2019objectnet,hendrycks2021many,hendrycks2021natural,dong2022viewfool}. 
To fully understand the effectiveness of existing algorithms and measure actual progress in the field, it is indispensable to comprehensively and correctly benchmark the  robustness of models under diverse settings \citep{carlini2019evaluating,dong2020benchmarking,croce2021robustbench}. It can also be an impetus for the development of more robust models. 

However, the research on model robustness is often faced with an \emph{arms race} between attacks and defenses, i.e., a defense method robust to existing attacks can be further evaded by new attacks, and vice versa \citep{Athalye2018Obfuscated,carlini2019evaluating,tramer2020adaptive},  
making robustness evaluation particularly challenging. 
Various works have been devoted to building robustness benchmarks
\citep{ling2019deepsec,dong2020benchmarking,croce2021robustbench,tang2021robustart}, which facilitate fair comparisons of different models. However, the existing benchmarks are not comprehensive enough. First, some of the benchmarks can hardly keep up with the state-of-the-art due to the rapid development of deep learning models, rendering the benchmarking results outdated. 
Second, most benchmarks mainly focus on adversarial robustness but rarely study robustness to natural distribution shifts and adversarial examples in tandem. 
As a result, they cannot disclose the relationship between different aspects of model robustness.

Moreover, the evaluation metrics lie in the core of robustness benchmarks. The existing benchmarks primarily adopt point-wise evaluation metrics to compare the robustness of models, which cannot comprehensively demonstrate their performance. For example, \cite{tang2021robustart} show that Transformers are more robust than CNNs against adversarial attacks, while \cite{mahmood2021robustness} observe that Transformers provide no additional robustness over CNNs. The conflicting results are due to the different noise budgets adopted in their benchmarks. Therefore, the point-wise evaluation metrics are insufficient to provide a global understanding of model robustness. To address this problem, our previous work \citep{dong2020benchmarking} proposes robustness curves as fair-minded evaluation metrics that can help to thoroughly compare the robustness of models at different noise levels. 

In this paper, 
we establish a comprehensive and rigorous benchmark called \textbf{ARES-Bench}\footnote{\textbf{ARES-Bench} is named after the platform called \textbf{Adversarial Robustness Evaluation for Safety (ARES)}.} to evaluate model robustness on the image classification task. Our benchmark evaluates both \emph{natural robustness} with 4 real-world and 3 synthesized OOD datasets, and \emph{adversarial robustness} with various white-box and black-box attack methods. We systematically study 55 models on ImageNet \citep{russakovsky2015imagenet} with diverse network architectures, including typical CNNs and Transformers, and different learning algorithms, including normal supervised training, pre-training on large-scale datasets \citep{dosovitskiy2021image}, self-supervised learning (SSL) \citep{chen2021empirical,he2022masked}, and adversarial training (AT) \citep{Madry2017Towards}.
Using robustness curves as the evaluation criteria, we conduct extensive experiments, based on which we draw some important findings. 
First, given a particular architecture, there is a trade-off between adversarial and natural robustness. Specifically, AT degrades natural robustness for most OOD datasets although the adversarial robustness is improved. We find that although AT learns robust features that are more shape-biased and aligned with humans \citep{tsipras2018robustness,zhang2019interpreting,ilyas2019adversarial}, these features generalize poorly to real-world distribution shifts.
Second, AT on Transformers performs much better than CNNs. Among the architectures we study, the optimal one is the Swin Transformer \citep{liu2021swin}, on the basis of which AT achieves more than $60\%$ robustness against the $\ell_\infty$-norm bounded perturbations of $4/255$. 
The main advantage of Swin Transformer is the hierarchical architecture with the self-attention mechanism, which allows AT to concentrate more on global features and benefit the adversarial robustness. 
Third, pre-training on large-scale datasets (e.g., ImageNet-21K) and by SSL significantly improves natural robustness. For AT, a pre-trained model on larger datasets is a better initialization than a random model, which speeds up AT. 
More analyses and discussions can be found in Sec.~\ref{sec4}.

Based on ARES-Bench, we further provide two case studies. First, we examine the effects of a wide range of tricks (e.g., data augmentation, regularization, weight averaging, pre-training, etc.) in large-scale AT on ImageNet. Most of these tricks prevent overfitting to the training data, which is a severe problem in AT. By performing ablation studies of these tricks, we obtain an optimized setting to achieve the state-of-the-art adversarial robustness compared with the existing methods \citep{xie2019feature,salman2020adversarially,debenedetti2022light}. Second, to explain why one model is more adversarially robust than another, we provide a frequency analysis to exhibit the frequency bias of the robust models. We find that AT models have lower frequency bias than normally trained models, suggesting AT uses more low-frequency or shape-biased features for prediction.

In summary, our main contributions include:
\begin{itemize}
    \item We establish ARES-Bench, a comprehensive benchmark to evaluate natural and adversarial robustness of image classifiers. Using robustness curves, we systematically investigate the robustness of 55 models on ImageNet with different architectures and learning algorithms.
    \item Based on the large-scale experiments, we have obtained many insightful findings, that reveal the inherent relationship between natural and adversarial robustness while providing a deeper understanding of robustness of different architectures and learning methods.
    \item Based on our benchmark, we analyze the training tricks in large-scale adversarial training on ImageNet and achieve the new state-of-the-art robustness. A frequency analysis is provided to exhibit the frequency bias of adversarially robust models.
    \item We make our benchmark available at \url{https://ml.cs.tsinghua.edu.cn/ares-bench}, which contains the leaderboards under various settings. We also open-source the robustness platform ARES 
(\url{https://github.com/thu-ml/ares}) and the collection of models in our benchmark.
\end{itemize}
We hope our benchmark, platform, robust models, and detailed analyses can be helpful for future research.

\section{Related Work}\label{sec2}

\begin{figure*}[t]
  \centering
\includegraphics[width=0.9\linewidth]{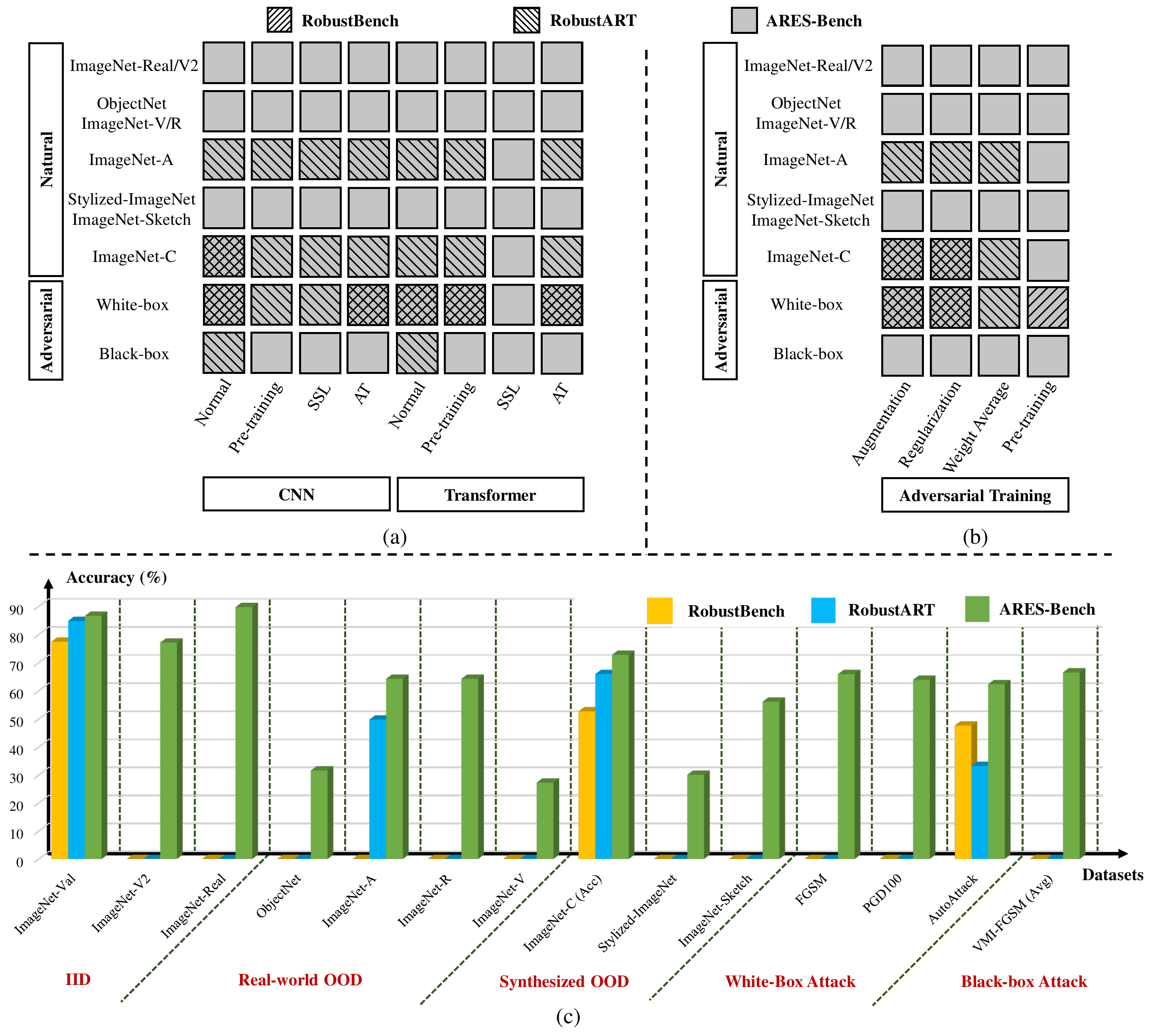}
  \caption{A comparison between our benchmark ARES-Bench and two existing benchmarks RobustBench \citep{croce2021robustbench} and RobustART \citep{tang2021robustart}. \textbf{(a)} shows the different types of models evaluated with various datasets and attacks in each benchmark. \textbf{(b)} shows the training tricks in adversarial training evaluated with various datasets and attacks in each benchmark. \textbf{(c)} shows the best results under each dataset and attack of the models in each benchmark. Compared with the existing benchmarks, our benchmark evaluates diverse and top-performing models under more OOD datasets and adversarial attacks. Therefore, our benchmark is more comprehensive and represents the state-of-the-art performance.
  }
  \label{fig:fig1}
\end{figure*}

\subsection{Deep Learning Robustness}

It has been widely demonstrated that deep learning models have poor robustness. \cite{szegedy2013intriguing} first show that deep neural networks are vulnerable to adversarial examples, which are generated by adding imperceptible perturbations to natural examples but significantly affect the model predictions. 
A number of works have been devoted to studying adversarial robustness. On the one hand, many adversarial attacks \citep{goodfellow2014explaining,kurakin2016adversarial,carlini2019evaluating,dong2018boosting,xie2019improving,croce2020reliable} have been proposed to improve the effectiveness and efficiency of generating adversarial examples under white-box and black-box settings. 
Adversarial attacks can serve as an important surrogate to identify the weaknesses and evaluate the robustness of deep learning models. On the other hand, numerous adversarial defense methods \citep{Kurakin2017Adversarial,tramer2017ensemble,Madry2017Towards,Guo2017Countering,Xie2018Mitigating,Wong2018Provable,liao2017Defense,cohen2019certified,zhang2019theoretically,pang2020bag} have been developed to improve model robustness. Among  the existing defenses, adversarial training (AT) is arguably the most effective technique \citep{Athalye2018Obfuscated,dong2020benchmarking}, in which the network is trained on the adversarial examples generated by attacks.

Besides adversarial robustness, deep learning models are also susceptible to natural distribution shifts. For example, \cite{engstrom2019exploring} demonstrate that an image translation or rotation is sufficient to mislead a target model. \cite{hendrycks2018benchmarking} introduce ImageNet-C/P to benchmark model robustness to common corruptions. \cite{geirhos2018imagenet} show that ImageNet classifiers are biased towards texture rather than shape. Other works \citep{barbu2019objectnet,hendrycks2021many,hendrycks2021natural,dong2022viewfool} have collected real-world datasets to 
evaluate out-of-distribution generalization performance. For natural robustness, there are also techniques to improve robustness
\citep{hendrycks2019augmix,hendrycks2021many,xie2020adversarial}.

\subsection{Robustness Benchmarks}

Since an increasing number of deep learning methods and robustness improvement techniques have been proposed, a comprehensive benchmark of model robustness is vital to understand their effectiveness and keep up with the state-of-the-art. 
Many works have developed robustness platforms that implement popular attacks for evaluating the robustness, including CleverHans \citep{papernot2016technical}, Foolbox \citep{rauber2017foolbox}, ART \citep{nicolae2018adversarial}, etc. But these platforms do not include the latest state-of-the-art 
models and do not provide benchmarking results.
Some robustness benchmarks have been further established \citep{ling2019deepsec,dong2020benchmarking,croce2021robustbench,tang2021robustart}. Nevertheless, most benchmarks focus on adversarial robustness rather than natural robustness, thus they are unable to reveal the inherent relationships between them.
For example, the widely used RobustBench \citep{croce2021robustbench} mainly evaluates adversarial robustness of adversarially trained models by AutoAttack \citep{croce2020reliable}.
The most relevant work to ours is RobustART \citep{tang2021robustart}, which benchmarks 
robustness
of dozens of network architectures and training techniques under various adversarial attacks and out-of-distribution datasets. The main limitation of RobustART is that it adopts aligned training settings to understand the robustness of each component, but this leads to inferior performance in the benchmark (i.e., it does not contain state-of-the-art models). It also incorporates insufficient OOD datasets. 

Compared with the existing benchmarks, ours is more comprehensive in the following aspects.
First, our benchmark integrates 3 IID and 7 OOD datasets to better evaluate natural robustness, as shown in Fig.~\ref{fig:fig1}(a) and (b).
Second, we evaluate a wide range of top-performing models, some of which are trained by ourselves through extensive ablation studies, leading to better results of the models in our benchmark compared with those in other benchmarks, as illustrated in Fig.~\ref{fig:fig1}(c).
Third, we adopt robustness curves as the evaluation criteria, which can fully show the model performance under varying noise severities. Fourth, we provide a frequency analysis to explore the frequency bias of adversarially trained models and explain the model behaviors. 

\section{Benchmark Design}\label{sec3}

\begin{table*}[t]\footnotesize
\caption{Evaluation datasets of robustness in ARES-Bench. We include 3 IID datasets and 7 OOD datasets to evaluate natural robustness. We adopt various white-box and transfer-based black-box attacks to evaluate adversarial robustness.
}
\label{tab:datasets}
\begin{center}
\begin{minipage}{0.9\textwidth}
\begin{tabular*}{\textwidth}{@{\extracolsep{\fill}}llllc@{\extracolsep{\fill}}}
\toprule%
Robustness & Distribution Shift & Source & Datasets \\
\midrule
\multirow{4}*{Natural} & IID & Real-world & ImageNet-Val, ImageNet-V2, ImageNet-Real \\
\cmidrule{2-4}
& \multirow{3}*{OOD} & Real-world & ObjectNet,  ImageNet-A, ImageNet-R, ImageNet-V \\
\cmidrule{3-4}
&& Synthesized & ImageNet-C, Stylized-ImageNet, ImageNet-Sketch \\
\midrule
\multirow{1}*{Adversarial}
& Adversarial & Adversary & White-box attacks, Transfer-based attacks \\
\botrule
\end{tabular*}
\end{minipage}
\end{center}
\end{table*}

In this paper, we establish \textbf{ARES-Bench}, a comprehensive benchmark for evaluating the robustness on image classification tasks. As two important aspects of model robustness, we investigate the relationship between natural and adversarial robustness using a variety of OOD datasets and adversarial attacks. For image classification models in the evaluation, we consider 55 models on ImageNet covering two mainstream types of network architectures: 
CNNs and Transformers, and four training paradigms that are closely related to robustness, including 
normal supervised training, pre-training on large-scale datasets, self-supervised learning (SSL), and adversarial training (AT). We systematically evaluate the robustness of these models using robustness curves. Below, we introduce the evaluation datasets and metrics in Sec.~\ref{sec:3-1}, and the model zoo in Sec.~\ref{sec:3-2}.

\subsection{Evaluation Datasets and Metrics}\label{sec:3-1}
In this section, we mainly introduce the evaluation datasets and adversarial attacks in our benchmark for natural robustness and adversarial robustness.

\subsubsection{Natural Robustness}
Natural noise comes from various sources in the real world, such as weather changes, sensor damage, object deformation, etc., which is unavoidable and destructive to deep learning models. 
To comprehensively evaluate the natural robustness of image classification models under diverse kinds of noises, 
our benchmark includes 3 independent and identically 
distributed (IID) datasets (\emph{ImageNet validation set}, 
\emph{ImageNet-V2} \citep{recht2019imagenet} and \emph{ImageNet-Real} \citep{beyer2020we}), 
4 OOD datasets collected in the real world (\emph{ObjectNet} \citep{barbu2019objectnet}, \emph{ImageNet-A} \citep{hendrycks2021natural}, \emph{ImageNet-R} \citep{hendrycks2021many} and \emph{ImageNet-V} 
\citep{dong2022viewfool}), and 3 algorithmically synthesized OOD datasets (\emph{ImageNet-C} \citep{hendrycks2018benchmarking}, \emph{Stylized-ImageNet} \citep{geirhos2018imagenet} and \emph{ImageNet-Sketch} \citep{wang2019learning}), as shown in Table~\ref{tab:datasets}.
These datasets are compatible with the ImageNet classifiers and commonly adopted to evaluate their natural robustness, while they focus on different aspects, such as viewpoint changes, common corruptions, style transfer. An integration of these datasets can thoroughly show the model behavior under various distribution shifts. Most of the datasets utilize classification accuracy as the evaluation metric, except for ImageNet-C.

\textbf{ImageNet validation} \citep[\textbf{IN-Val},][]{russakovsky2015imagenet} is by default adopted to evaluate the clean accuracy of ImageNet classifiers. It contains 50,000 images belonging to 1,000 categories.

\textbf{ImageNet-V2} \citep[\textbf{IN-V2},][]{recht2019imagenet} is a new test dataset for the ImageNet benchmark, which is used to evaluate the generalization ability of image classifiers. It contains three test sets with 10,000 new images each. We only use the subset of matched-frequency-format.

\textbf{ImageNet-Real} \citep[\textbf{IN-Real},][]{beyer2020we} adjusts the labels in ImageNet, which considers multiple less important labels of an image. It is used to test whether the model overfits to the idiosyncrasies of the original labeling procedure. 

\textbf{ObjectNet} \citep[\textbf{ON},][]{barbu2019objectnet} is a real-world test dataset containing 50,000 images, in which object backgrounds, rotations, and imaging viewpoints are random. It is used to test robustness under background and viewpoint changes. 

\textbf{ImageNet-A} \citep[\textbf{IN-A},][]{hendrycks2021natural} is a collection of hard-to-classify samples against the ResNet50~\citep{he2015deep} model. The samples in ImageNet-A are real-world and unmodified with limited spurious cues. 

\textbf{ImageNet-R} \citep[\textbf{IN-R},][]{hendrycks2021many} contains real-world images with changes in image style, including art, cartoons, etc. It has renditions of 200 ImageNet classes and totally 30,000 images, which can be used to evaluate robustness to real-world style transferred images.

\textbf{ImageNet-V} \citep[\textbf{IN-V},][]{dong2022viewfool} is a recently proposed dataset to evaluate viewpoint robustness of image classifiers. It contains 10,000 images of 3D objects collected from adversarial viewpoints generated by an attack method.

\textbf{ImageNet-C} \citep[\textbf{IN-C},][]{hendrycks2018benchmarking} consists of 15 types of synthesized corruptions ranging from noise, blur, weather, to digital ones. Each corruption is associated with 5 severities. ImageNet-C has been widely used to evaluate corruption robustness of image classification models. Commonly, the metric of ImageNet-C is the normalized mean corruption error (mCE).

We let $f_{j,k}(\cdot)$ denote the function of the $j$-th corruption type under the $k$-th severity. For the original dataset $\mathcal{D}=\{(\bm{x}_i, y_i)\}_{i=1}^N$ with $N$ samples, the corruption dataset is constructed by applying every corruption function $f_{j,k}(\cdot)$ to the original dataset. Then, the corruption error of a classifier $\mathcal{C}$ under the $j$-th corruption is defined as
\begin{equation}
    \mathrm{CE}(\mathcal{C}, j)=\frac{1}{NK}\sum_{i=1}^N \sum_{k=1}^K\bm{1}(\mathcal{C}(f_{j,k}(\bm{x}_i)) \neq y_i),
\end{equation}
where $\bm{1}(\cdot)$ is an indicator function. When it comes to multiple corruptions, the mean corruption error (mCE) can be defined as
\begin{equation}
    \mathrm{mCE}(\mathcal{C})=\frac{1}{M}\sum_{j=1}^M \frac{\mathrm{CE}(\mathcal{C}, j)}{\mathrm{CE}(\mathrm{AlexNet}, j)},
\end{equation}
where $M$ is the total number of corruptions and the performance of AlexNet is used as the baseline.

Apart from the mCE metric, we also introduce a robustness curve of \emph{classification accuracy vs. severity} of the corruptions, which can be used to study robustness under different severities.

\textbf{Stylized-ImageNet} \citep[\textbf{SIN},][]{geirhos2018imagenet}
is a stylized version of ImageNet, which is generated by style transfer and can be used to measure shape bias of models.

\textbf{ImageNet-Sketch} \citep[\textbf{IN-Sketch},][]{wang2019learning} contains 50,000 sketch-like images of the 1,000 ImageNet classes. All images are within the ``black and white'' color scheme.

\subsubsection{Adversarial Robustness}

Adversarial robustness is an important aspect of model robustness in the presence of adversaries. 
To measure adversarial robustness, it is typical to adopt adversarial attacks as surrogates, which craft the worst-case adversarial examples under a specific threat model.
To make the adversarial example $\bm{x}^{adv}$ visually indistinguishable from the original example $\bm{x}$, a norm bound is usually adopted to restrict the perturbation as $\|\bm{x}^{adv}-\bm{x}\|_p\leq \epsilon$, in which $\epsilon$ is the perturbation budget. 
An adversarial example can be generated by solving
\begin{equation}\label{eq:adv}
\bm{x}^{adv}=\mathop{\arg \max} \limits_{\hat{\bm{x}}:\|\hat{\bm{x}}-\bm{x}\|_p\leq\epsilon} \mathcal{L}(\hat{\bm{x}}, y),
\end{equation}
where $\mathcal{L}$ is a classification loss (e.g., cross-entropy loss, C\&W loss \citep{carlini2017towards}, etc.).

To solve the optimization problem \eqref{eq:adv}, various gradient-based attack methods have been proposed \citep{goodfellow2014explaining,kurakin2016adversarial,carlini2017towards}. These methods calculate the gradient of the loss function w.r.t. input,  which are known as \emph{white-box attacks}. Besides, researchers have found that adversarial examples exhibit good cross-model transferability \citep{liu2016delving}, making \emph{black-box attacks} feasible.

In our benchmark, we evaluate adversarial robustness with 3 white-box attacks, including \emph{Fast Gradient Sign Method (FGSM)} \citep{goodfellow2014explaining}, \emph{Projected Gradient Descent (PGD)} \citep{Madry2017Towards} and \emph{AutoAttack} \citep{croce2020reliable}, and 5 black-box attacks, including \emph{Momentum Iterative Method (MIM)} \citep{dong2018boosting}, \emph{Diversity Input Method (DIM)} \citep{xie2019improving}, \emph{Translation Invariant Method (TIM)} \citep{dong2019evading}, \emph{Scale-Invariant Nesterov Iterative Method (SI-NI-FGSM)} \citep{lin2019nesterov} and \emph{Variance Tuning Method (VMI-FGSM)} \citep{wang2021enhancing}. We include the representative and state-of-the-art attack methods under the white-box and black-box settings, e.g., AutoAttack for white-box attacks and VMI-FGSM for black-box attacks, 
which facilitate a more correct evaluation of adversarial robustness.
We consider both $\ell_\infty$-norm and $\ell_2$-norm perturbations.
The detailed settings of these attacks are shown in Appendix A.

For the evaluation metrics, we follow our previous work \citep{dong2020benchmarking} to adopt the robustness curves. 
In this work, we consider the curve of \emph{accuracy vs. perturbation budget}, which shows the robustness of models under different perturbation budgets, to provide a global understanding of robustness. 
To efficiently obtain a robustness curve, we first perform a binary search for a sample to find the minimum perturbation budget that can lead to the misclassification of the crafted adversarial example. Then, we compute the percentage of data samples whose minimum perturbations are smaller than each $\epsilon$ to plot the curve.
For black-box attacks, we show the transferability heatmap of all models.

\begin{table*}[t]\footnotesize
\caption{Image classification models in our benchmark. We include typical architectures of CNNs and Transformers with different sizes. The models are trained by normal supervised learning, pre-training on ImageNet-21K, self-supervised learning (SSL) and adversarial training (AT). We also include existing adversarially robust models from RobustBench \citep[RB,][]{croce2021robustbench}, Robust Library \citep[RL,][]{robustness}, and feature denoising \citep[FD,][]{xie2019feature}.}
\label{tab:arch}
\begin{center}
\begin{minipage}{0.9\textwidth}
\begin{tabular*}{\textwidth}{@{\extracolsep{\fill}}llllccccc@{\extracolsep{\fill}}}
\toprule%
\multicolumn{3}{c}{\multirow{3}*{Model Architecture}} & \multirow{3}*{Size} & \multicolumn{5}{c}{Training Paradigm} \\
\cmidrule{5-9}
&&&&Normal&Pre-training&SSL&AT&Others\\
\midrule
\multirow{16}*{CNN}
& \multirow{3}*{VGG} & VGG13 &127M& \checkmark \\
&& VGG16 &131M& \checkmark \\
&& VGG19 &137M& \checkmark \\
\cmidrule{2-9}
& \multirow{4}*{ResNet} & ResNet50 &24M& \checkmark&&\checkmark&\checkmark&RB,RL\\
&& ResNet101 &42M& \checkmark&&&\checkmark \\
&& ResNet152 &57M& \checkmark&&&\checkmark&FD\\
&& Wide-ResNet50 &66M& \checkmark&&&\checkmark&RB \\
\cmidrule{2-9}
& \multirow{3}*{DenseNet} & DenseNet121 &8M& \checkmark \\
&& DenseNet161 &27M& \checkmark \\
&& DenseNet201 &19M& \checkmark \\
\cmidrule{2-9}
& \multirow{3}*{ConvNext} & ConvNextS &48M& \checkmark&\checkmark&&\checkmark \\
&& ConvNextB &84M& \checkmark&\checkmark&&\checkmark \\
&& ConvNextL &189M& \checkmark&\checkmark&&\checkmark \\
\midrule
\multirow{15}*{Transformer}
& \multirow{3}*{ViT} & ViTS &21M& \checkmark&\checkmark&&\checkmark \\
&& ViTB &83M& \checkmark&\checkmark&\checkmark&\checkmark \\
&& ViTL &290M& \checkmark&\checkmark&\checkmark \\
\cmidrule{2-9}
& \multirow{3}*{XciT} & XciTS &45M& \checkmark& &&&RB \\
&& XciTM &80M& \checkmark &&&&RB\\
&& XciTL &180M& \checkmark &&&&RB\\
\cmidrule{2-9}
& \multirow{3}*{T2T} & T2T14 &20M& \checkmark \\
&& T2T19 &37M& \checkmark \\
&& T2T24 &61M& \checkmark \\
\cmidrule{2-9}
& \multirow{3}*{Swin} & SwinS &47M& \checkmark&\checkmark& &\checkmark \\
&& SwinB &84M& \checkmark&\checkmark&&\checkmark \\
&& SwinL &187M& &\checkmark&&\checkmark \\
\botrule
\end{tabular*}
\end{minipage}
\end{center}
\end{table*}

\subsubsection{Frequency Perspective}
\label{sec:3.2.3}

Adversarial training has been shown to effectively improve adversarial robustness by replacing the benign inputs with the adversarial examples, but this training paradigm remains opaque. An interpretability technique is necessary for researchers to understand the mechanism of AT and develop better architectures and training strategies. Generally, interpretability approaches highlight the important regions that affect the decision-making  \citep{selvaraju2017grad,cao2020analyzing}. However, the adversarial noise is overlapped with the benign image, making it difficult to find out the region of interest in the spatial domain. Inspired by the observation that adversarial perturbations are actually high-frequency signals \citep{maiya2021frequency, zhang2019adversarial}, we intend to find out the differences between AT models and the normally trained ones for their frequency responses.

From the perspective of Fourier analysis~\citep{howell2016principles}, signals can be decomposed into several basic trigonometric functions. Therefore, we can isolate the signals that contribute to the final decision to find out the frequency bias of the models. Specifically, we develop a low-pass filter to drop high-frequency signals. By gradually increasing the band-width of the low-pass filter, the minimum cutoff frequency $f_c$ for each sample is determined. The minimum cutoff frequency indicates how many trigonometric signals are sufficient for the correct classification. The ensemble of the minimum $f_c$ of all samples forms the final frequency bias of a model. To exhibit the ensemble of $f_c$, we introduce the \emph{accuracy vs. low-pass bandwidth} (ACC-LPB) curve. Besides, it is meaningless to examine $f_c$ of misclassified clean samples, so we can normalize the accuracy to 0-1 by dividing all the accuracy with the clean accuracy. Thus, we introduce a metric to measure the frequency bias as
\begin{equation}\label{eq:frenq}
    f_{bias}=\frac{1}{N'} \sum^{N'}_{i=1} \mathop{\min} f_c(\bm{x}_i),
\end{equation}
where $N'$ is the number of correctly classified 
samples, 
$f_c(\bm{x}_i)$ is the low pass bandwidth that can maintain the correct classification. This metric is positive linear correlated to the area ratio of the part above the ACC-LPB curve.

\subsection{Model Zoo}\label{sec:3-2}

In this section, we introduce the image classification models evaluated in our benchmark. Recently, Vision Transformer \citep{dosovitskiy2021image} has become the prevalent model architecture for image recognition based on the long-range self-attention mechanism. To fully understand their robustness compared with convolutional neural networks (CNNs), we include typical networks of CNNs and Transformers. Besides normal supervised training on ImageNet-1K, we further consider other typical training paradigms, including pre-training on large-scale datasets (e.g., ImageNet-21K), self-supervised learning (SSL), and adversarial training (AT), to facilitate a deeper understanding of their effects on model robustness because they are widely studied in the literature. Therefore, we include a total number of 55 models in the evaluation, as shown in Table~\ref{tab:arch}.
Besides these models, we further study training tricks in large-scale AT on ImageNet, as detailed in Sec.~\ref{sec:3.2.2}.

\subsubsection{Architectures and Training Paradigms}\label{sec:3-2-1}

\textbf{Network architectures.} 
To comprehensively understand the robustness of Transformers compared with CNNs, we include several typical and advanced architectures. For CNNs, we consider \emph{VGG} \citep{simonyan2014very}, \emph{ResNet} \citep{he2015deep}, \emph{DenseNet} \citep{huang2017densely}, and \emph{ConvNext} \citep{liu2022convnet}. The first three are famous architectures in the field while the last one is recently developed to achieve comparable performance with Transformers. For  Transformer models, we consider \emph{ViT} \citep{dosovitskiy2021image}, \emph{XciT} \citep{el2021xcit}, \emph{T2T} \citep{Yuan_2021_ICCV}, and \emph{Swin Transformer} \citep{liu2021swin}.
For these architectures, we also consider different model sizes, as shown in Table~\ref{tab:arch}.
We collect most of the normally trained models on ImageNet-1K of these architectures. Besides, we also study other training paradigms, as illustrated below.

\textbf{Pre-training on large-scale datasets.} This is a common strategy to prevent overfitting and improve the generalization performance \citep{dosovitskiy2021image,liu2022convnet}. However, its effects on model robustness are rarely explored. To understand its effects, we consider ImageNet-21K as the pre-training dataset, which contains 21K classes and 14M images \citep{deng2009imagenet}. In our benchmark, we include ConvNext, ViT, and Swin based on the released models. 

\textbf{Self-supervised learning (SSL)}. SSL is an effective method of learning discriminative representations from unlabeled data based on preset tasks. The paradigm of pre-training by SSL and fine-tuning on downstream tasks has become a prevailing approach. Despite the effectiveness, the study on the effects of SSL on model robustness is limited. Studying this problem is essential to understand the behavior of SSL when deployed in security-sensitive applications. In our benchmark, we consider two SSL methods, including MOCOv3 \citep{chen2021empirical} and MAE \citep{he2022masked}, as they are among the most representative SSL methods and achieve superior performance. MOCOv3 adopts a ResNet50 backbone and MAE is based on the ViT architecture.

\textbf{Adversarial training (AT).} AT augments training data with adversarial examples \citep{goodfellow2014explaining,Madry2017Towards}, which is shown to be the most effective technique of improving adversarial robustness \citep{Athalye2018Obfuscated,dong2020benchmarking,croce2020reliable}. Formally, AT can be formulated as a minimax optimization problem \citep{Madry2017Towards}:
\begin{equation}
    \min \limits_{\bm{w}} \mathbb{E}_{(\bm{x},y) \sim \mathcal{D}}\max \limits_{\hat{\bm{x}}:\|\hat{\bm{x}}-\bm{x}\|_p\leq\epsilon}\mathcal{L}(\hat{\bm{x}},y;\bm{w}),
\end{equation}
where $\bm{w}$ denotes the weights of a classifier.
The inner maximization can be solved by PGD \citep{Madry2017Towards}. In our benchmark, we perform AT with the  $\ell_\infty$-norm bounded perturbations of $\epsilon=4/255$. To accelerate training on ImageNet-1K, we adopt the PGD-3 adversary with the step size $2\epsilon/3$. Due to the heavy cost of training, we do not run AT for all architectures, but only consider ResNet, ConvNext, ViT, and Swin to cover the most widely used and state-of-the-art networks of CNNs and Transformers. 
The training settings of AT are similar to normal training, as detailed in Appendix A. We also introduce the training tricks used in AT in Sec.~\ref{sec:3.2.2}.

For comparison, we also incorporate the existing state-of-the-art adversarially robust models according to RobustBench \citep{croce2021robustbench}, including a ResNet50 and a Wide-ResNet50 from \cite{salman2020adversarially}, three XciT models from \cite{debenedetti2022light}, a ResNet50 model from the Robust Library \citep[RL,][]{robustness}, and a ResNet152 based on Feature Denoising \citep[FD,][]{xie2019feature}.

\subsubsection{Training Tricks in AT}
\label{sec:3.2.2}
\begin{table}[t]
\begin{center}
\begin{minipage}{215pt}
\caption{Training tricks in adversarial training.}
\label{tab:strategy}%
\begin{tabular}{@{}lc@{}}
\toprule
Category & Method \\
\midrule
Data Augmentation & Mixup, RandAugment \\
Regularization & Weight Decay, Label Smoothing \\
Weight Averaging & EMA \\
Pre-training & 21K-pre-training, SimMIM, CLIP \\
\botrule
\end{tabular}
\end{minipage}
\end{center}
\end{table}

Recent works have found that the training tricks (e.g., weight decay, label smoothing, weight averaging) play an important role in AT \citep{pang2020bag,gowal2020uncovering}. However, previous works mainly study the tricks in AT on smaller datasets (e.g., CIFAR-10), but do not explore large-scale AT on ImageNet. Only \cite{debenedetti2022light} provide a recipe of training robust Vision Transformers on ImageNet.
As adversarial training is an important method evaluated in our benchmark, we put a special focus on various training tricks in large-scale AT to thoroughly understand their effectiveness and provide a guideline for training robust models on ImageNet.

Based on previous studies \citep{pang2020bag,gowal2020uncovering}, we only concentrate on four categories of training tricks that have a significant impact on adversarial robustness, including data augmentation, regularization, weight averaging, and pre-training. Table~\ref{tab:strategy} shows the studied methods of each category, most of which have also been integrated into the recent normally trained models such as ConvNext and Swin. Below, we detail these methods.

\textbf{Data augmentation (DA).} DA techniques are commonly adopted to prevent overfitting and improve the generalization performance of deep learning models. It has also been shown that DA can improve adversarial robustness \citep{rebuffi2021data}. In this work, we study Mixup \citep{zhang2017mixup} and RandAugment \citep{cubuk2020randaugment} as two typical DA techniques in AT. Specifically, Mixup blends two random samples and their labels with a certain ratio, which can be expressed as
\begin{equation}
\tilde{\bm{x}}=\lambda\bm{x}_i+(1-\lambda)\bm{x}_j;\;
\tilde{\bm{y}}=\lambda\bm{y}_i+(1-\lambda)\bm{y}_j,
\end{equation}
where $\bm{x}_i,\bm{x}_j$ are two samples, $\bm{y}_i,\bm{y}_j$ are their corresponding labels. $\tilde{\bm{x}}, \tilde{\bm{y}}$ are the generated samples for training.
RandAugment performs random transformations (e.g., auto-contrast, shear transformation, etc.) of images. It reduces the space for searching the augmentation policies and  the regularization strength can be tailored to different models and dataset sizes.

\textbf{Regularization.} As shown in \cite{pang2020bag}, weight decay and label smoothing are two important regularization methods that can significantly affect the performance of AT. A proper value of weight decay can enlarge the margin of a sample from the decision boundary by imposing a $\ell_2$ regularization on model weights. Label smoothing introduces uncertainty to the one-hot labels, which can combat with label noise in AT to avoid robust overfitting \citep{dong2022exploring}.

\textbf{Weight averaging (WA).} WA is widely used in normal training to find flatter optima and lead to better generalization \citep{izmailov2018averaging}. It has been shown that WA has a significant robustness improvement \citep{gowal2020uncovering}.
For WA, we adopt an exponential moving average \citep[EMA,][]{bolme2009average} of model weights as
\begin{equation}
    \tilde{\bm{w}}_t=\beta \cdot \tilde{\bm{w}}_{t-1} + (1-\beta) \cdot \bm{w}_t,
\end{equation}
where $\tilde{\bm{w}}_t$ denotes the average model weights at the $t$-th training step, $\bm{w}_t$ denotes the current model weights, and $\beta$ is a hyperparameter commonly set between $0.9\sim0.999$.

\textbf{Pre-training.} It has been shown that pre-training on large-scale datasets can improve model robustness \citep{hendrycks2019using}. Pre-trained models by SSL can also be adversarially fine-tuned towards adversarial robustness on downstream tasks \citep{chen2020adversarial,dong2021should}. Thus, we further study pre-training for large-scale AT. Different from the previous works, our purpose is to examine whether the pre-trained models can serve as better initializations for AT to accelerate training, thus we do not perform adversarial pre-training but adopt the pre-trained models given by different methods.
Specifically, we consider three widely studied types of pre-training: pre-training on large-scale datasets, masked image modeling, and vision-language pre-training. For pre-training on large-scale datasets, we adopt ImageNet-21K as the pre-training dataset (denoted as \textbf{21K-pre-training}). We think that more data can prevent models from overfitting and improve the robustness. For masked image modeling, we adopt \textbf{SimMIM} \citep{xie2022simmim}. Although similar to MAE, we choose SimMIM because it adopts the backbone of Swin Transformer, which performs better than ViT adopted in MAE, as shown in the experiment. 
For vision-language pre-training, we focus on the famous \textbf{CLIP} model \citep{radford2021learning}, which performs contrastive learning on images and the corresponding texts. We only adopt the image encoder from CLIP for adversarial fine-tuning.

\renewcommand{\arraystretch}{1.2}
\begin{table*}[t]\footnotesize
\caption{The natural robustness of CNN-based models, including VGG, ResNet, DenseNet and ConvNext. We show the classification accuracy (\%) on these datasets and $1-\mathrm{mCE}$ on ImageNet-C for consistent comparison (higher is better). We mark the best result within each architecture in \textbf{bold}, and mark the overall best result in {\color{red} \textbf{red}}.
}
\label{tab:ood-cnn}
\begin{center}
\begin{minipage}{0.99\textwidth}
 \setlength{\tabcolsep}{5pt}{
\begin{tabular*}{\textwidth}{@{\extracolsep{\fill}}ll ccc cccc cccc@{\extracolsep{\fill}}}
\toprule%
\multirow{3}*{Architecture}& \multirow{3}*{Method} & \multicolumn{3}{c}{IID} & \multicolumn{4}{c}{Real-world OOD} & \multicolumn{3}{c}{Synthesized OOD} & \multirow{3}*{Avg.} \\
\cmidrule{3-5}\cmidrule{6-9}\cmidrule{10-12}
& & IN-Val & IN-V2 & IN-Real & ON & IN-A & IN-R & IN-V & IN-C & SIN & IN-Sketch \\
\midrule
VGG13 & Normal & 69.6 & 57.3 & 77.0 & 10.4 & 2.2 & 25.5 & 9.1 & 6.0 & 2.8 & 15.4 & 27.5 \\
\cmidrule{2-13}
VGG16 & Normal & 71.4 & 58.6 & 78.8 & 12.5 & 2.8 & 27.0 & 10.1 & 9.7 & 3.0 & 16.9 & 29.1 \\
\cmidrule{2-13}
VGG19 & Normal & 72.2 & 59.7 & 79.2 & 13.4 & 2.4 & 27.6 & 9.6 & 11.1 & 2.8 & 17.2 & 29.5 \\
\midrule
\multirow{5}*{ResNet50} & Normal & \bf75.9 & \bf63.3 & \bf82.6 & \bf17.9 & 0.1 & 35.4 & \bf12.5 & 23.3 & 7.4 & 23.0 & 34.1 \\
 & MOCOv3 & 74.6 & 62.0 & 81.3 & 14.3 & \bf4.0 & 36.6 & 9.8 & \bf26.1 & 8.1 & \bf25.2 & \bf34.2 \\
 & AT & 66.1 & 53.0 & 73.7 & 9.1 & 2.6 & \bf40.3 & 7.5 & 15.3 & \bf13.2 & 21.8 & 30.2 \\
 & RB & 64.2 & 51.4 & 71.9 & 8.5 & 2.3 & 38.3 & 7.1 & 14.9 & 11.7 & 20.9 & 29.1 \\
 & RL & 62.9 & 50.2 & 71.0 & 8.0 & 2.0 & 39.4 & 7.2 & 15.0 & 12.5 & 20.8 & 28.9 \\
 \cmidrule{2-13}
\multirow{2}*{ResNet101} & Normal & \bf77.3 & \bf65.7 & \bf83.6 & \bf19.1 & \bf5.1 & 38.7 & \bf13.6 & \bf29.6 & 9.1 & \bf26.4 & \bf36.8 \\
 & AT & 70.4 & 57.7 & 78.3 & 11.8 & 3.7 & \bf42.9 & 8.4 & 20.6 & \bf15.2 & 25.1 & 33.4 \\
 \cmidrule{2-13}
\multirow{3}*{ResNet152} & Normal & \bf78.2 & \bf66.9 & \bf84.7 & \bf20.0 & \bf6.4 & 40.5 & \bf14.8 & \bf30.7 & 10.3 & 27.9 & \bf38.0 \\
 & AT & 72.4 & 60.0 & 79.7 & 13.0 & 5.4 & \bf47.2 & 9.4 & 24.4 & \bf16.7 & \bf29.9 & 35.8 \\
 & FD & 65.4 & 49.0 & 72.1 & 8.0 & 3.4 & 43.5 & 6.8 & 17.5 & 15.1 & 23.8 & 30.4 \\
 \cmidrule{2-13}
\multirow{3}*{Wide-ResNet50} & Normal & \bf81.5 & \bf70.5 & \bf86.6 & \bf22.9 & \bf16.3 & \bf44.2 & \bf16.1 & \bf40.4 & 12.0 & \bf32.5 & \bf42.3 \\
 & AT & 70.1 & 56.9 & 78.0 & 11.0 & 3.4 & 42.0 & 7.5 & 18.5 & 13.2 & 23.8 & 32.4 \\
 & RB & 68.8 & 55.8 & 76.4 & 10.6 & 3.3 & 41.8 & 8.2 & 19.5 & \bf13.6 & 23.7 & 32.2 \\
 \midrule
DenseNet121 & Normal & 74.7 & 62.6 & 81.7 & 15.8 & 2.6 & 36.8 & 12.3 & 26.6 & 8.0 & 23.8 & 34.5 \\
\cmidrule{2-13}
DenseNet161 & Normal & 77.4 & 65.8 & 83.8 & 17.8 & 4.7 & 39.6 & 14.3 & 33.6 & 11.3 & 28.1 & 37.6 \\
\cmidrule{2-13}
DenseNet201 & Normal & 77.3 & 65.2 & 83.5 & 17.9 & 4.1 & 40.3 & 13.4 & 31.6 & 10.9 & 27.3 & 37.1 \\
\midrule
\multirow{3}*{ConvNextS} & Normal & 83.2 & 72.5 & 88.0 & 25.3 & 31.3 & 49.6 & 17.8 & 50.5 & 21.8 & 37.1 & 47.7 \\
 & Pre-train & \bf84.6 & \bf74.7 & \bf89.1 & \bf28.7 & \bf44.8 & \bf57.6 & \bf23.2 & \bf56.3 & 19.2 & \bf43.6 & \bf52.2 \\
 &AT&76.1&63.8&82.8&15.8&10.5&53.7&11.0&36.5&\bf22.0&39.4&41.2\\
 \cmidrule{2-13}
\multirow{3}*{ConvNextB} & Normal & 83.8 & 73.7 & 88.2 & 26.3 & 36.6 & 51.3 & 19.9 & 53.2 & 21.5 & 38.2 & 49.3 \\
 & Pre-train & \bf85.8 & \bf76.0 & \bf89.6 & \bf30.5 & \bf54.6 & \bf62.0 & \bf24.5 & \bf59.6 & \bf25.1 & \bf48.8 & \bf55.7 \\
 &AT&77.1&64.9&83.8&16.5&12.2&56.8&11.5&38.8&23.6&43.6&42.9\\
 \cmidrule{2-13}
\multirow{3}*{ConvNextL} & Normal & 84.3 & 74.2 & 88.5 & 27.1 & 41.3 & 53.5 & 22.0 & 55.4 & 23.9 & 40.1 & 51.0 \\
 & Pre-train & {\color{red} \bf86.6} & {\color{red} \bf77.1} & {\color{red} \bf89.7} & {\color{red} \bf30.6} & {\color{red} \bf59.8} & {\color{red} \bf64.2} & {\color{red} \bf27.3} & {\color{red} \bf63.7} & 24.9 & {\color{red} \bf49.9} & {\color{red} \bf57.4}\\
 &AT&78.1&66.2&84.6&17.7&14.1&59.4&12.8&40.7&{\color{red} \bf25.2}&45.1&44.4\\
\botrule
\end{tabular*}
}
\end{minipage}
\end{center}
\end{table*}

\renewcommand{\arraystretch}{1.2}
\begin{table*}[t]\footnotesize
\begin{center}
\begin{minipage}{0.99\textwidth}
 \setlength{\tabcolsep}{5pt}
\caption{The natural robustness of Transformer-based models, including ViT, XciT, T2T and Swin. We show the classification accuracy (\%) on these datasets and $1-\mathrm{mCE}$ on ImageNet-C for consistent comparison (higher is better). We mark the best result within each architecture in \textbf{bold}, and mark the overall best result in {\color{red} \textbf{red}}.}
\label{tab:ood-transformer}
\begin{tabular*}{\textwidth}{@{\extracolsep{\fill}}ll ccc cccc cccc@{\extracolsep{\fill}}}
\toprule%
\multirow{3}*{Architecture}& \multirow{3}*{Method} & \multicolumn{3}{c}{IID} & \multicolumn{4}{c}{Real-world OOD} & \multicolumn{3}{c}{Synthesized OOD} & \multirow{3}*{Avg.} \\
\cmidrule{3-5}\cmidrule{6-9}\cmidrule{10-12}
& & IN-Val & IN-V2 & IN-Real & ON & IN-A & IN-R & IN-V & IN-C & SIN & IN-Sketch \\
\midrule
\multirow{3}*{ViTS} & Normal & 74.4 & 61.6 & 80.0 & 13.1 & 8.8 & 30.4 & 11.2 & 32.0 & 9.1 & 19.9 & 34.0 \\
 & Pre-train & \bf81.4 & \bf70.3 & \bf86.8 & \bf22.7 & \bf27.3 & \bf45.7 & \bf16.6 & \bf47.1 & 15.8 & \bf32.5 & \bf44.6 \\
 & AT & 70.2 & 57.3 & 77.9 & 11.5 & 6.1 & 46.0 & 8.5 & 27.8 & \bf16.8 & 29.8 & 35.2 \\
 \cmidrule{2-13}
\multirow{4}*{ViTB} & Normal & 75.8 & 61.6 & 80.9 & 13.2 & 11.4 & 32.8 & 13.3 & 34.3 & 10.9 & 23.7 & 35.8 \\
 & Pre-train & \bf84.6 & \bf73.9 & \bf88.8 & \bf27.4 & \bf44.5 & \bf56.8 & \bf19.4 & \bf57.5 & \bf22.6 & \bf43.0 & \bf51.9 \\
 & MAE & 83.6 & 73.1 & 88.1 & 24.9 & 37.7 & 49.8 & 18.2 & 49.4 & 20.2 & 36.4 & 48.1 \\
 & AT & 73.4 & 60.4 & 80.5 & 12.7 & 8.9 & 50.7 & 9.4 & 36.6 & 22.2 & 35.7 & 39.1 \\
 \cmidrule{2-13}
\multirow{3}*{ViTL} & Normal & 75.2 & 60.7 & 79.8 & 11.2 & 11.3 & 33.3 & 13.4 & 35.4 & 9.3 & 25.0 & 35.4 \\
 & Pre-train & \bf85.8 & \bf76.0 & \bf89.2 & \bf30.5 & \bf56.1 & {\color{red}\bf64.2} & \bf25.5 & {\color{red}\bf65.3} & {\color{red}\bf30.1} & {\color{red}\bf51.8} & {\color{red}\bf57.4} \\
 & MAE & 85.1 & 75.6 & 89.0 & 27.3 & 50.6 & 60.0 & 21.5 & 56.2 & 24.1 & 46.4 & 53.6 \\
 \midrule
\multirow{2}*{XciTS} & Normal & \bf82.4 & \bf71.5 & \bf86.8 & \bf23.7 & \bf31.3 & 45.0 & \bf17.0 & \bf50.1 & \bf19.5 & \bf32.9 & 
\bf46.0 \\
 & RB & 73.3 & 60.5 & 80.6 & 12.7 & 6.3 & \bf45.7 & 9.7 & 28.5 & 18.4 & 31.2 & 36.7 \\
 \cmidrule{2-13}
\multirow{2}*{XciTM} & Normal & \bf82.6 & \bf71.0 & \bf86.8 & \bf23.4 & \bf33.3 & 44.7 & \bf17.7 & \bf50.5 & \bf20.3 & \bf33.1 & \bf46.3 \\
 & RB & 74.1 & 61.7 & 81.3 & 13.6 & 7.0 & \bf47.1 & 9.5 & 30.2 & 19.7 & 32.6 & 37.7 \\
 \cmidrule{2-13}
\multirow{2}*{XciTL} & Normal & \bf83.0 & \bf72.0 & \bf86.9 & \bf23.7 & \bf36.2 & 46.2 & \bf17.9 & \bf50.2 & \bf20.4 & \bf34.4 & \bf47.1 \\
 & RB & 75.1 & 62.7 & 81.7 & 13.4 & 8.8 & \bf49.0 & 10.7 & 32.0 & 19.9 & \bf34.4 & 38.7\\
 \midrule
T2T14 & Normal & 81.6 & 70.9 & 86.8 & 22.3 & 24.1 & 44.7 & 16.7 & 46.8 & 17.7 & 32.2 & 44.4 \\
\cmidrule{2-13}
T2T19 & Normal & 82.3 & 71.6 & 87.2 & 23.2 & 29.0 & 47.3 & 18.0 & 50.2 & 20.9 & 34.4 & 46.4 \\
\cmidrule{2-13}
T2T24 & Normal & 82.4 & 71.7 & 87.2 & 22.9 & 29.7 & 47.9 & 18.0 & 52.0 & 20.8 & 35.1 & 46.8 \\
 \midrule
\multirow{3}*{SwinS} & Normal & 83.2 & 72.1 & 87.5 & 24.7 & 33.0 & 44.9 & 19.3 & 45.1 & 16.8 & 32.0 & 45.8 \\
 & Pre-train & \bf83.3 & \bf73.5 & \bf88.6 & \bf28.1 & \bf43.9 & \bf54.8 & \bf21.3 & \bf50.6 & 17.2 & \bf41.2 & \bf50.3 \\
 & AT & 75.8 & 63.3 & 82.6 & 15.3 & 10.6 & 52.5 & 10.8 & 37.1 & \bf21.1 & 37.1 & 40.6 \\
 \cmidrule{2-13}
\multirow{3}*{SwinB} & Normal & 83.4 & 72.3 & 87.6 & 25.5 & 35.8 & 46.6 & 20.2 & 45.6 & 17.9 & 32.4 & 46.7 \\
 & Pre-train & \bf85.1 & \bf75.2 & \bf89.1 & \bf28.8 & \bf51.8 & \bf59.1 & \bf22.7 & \bf56.4 & 19.6 & \bf45.1 & \bf53.3 \\
 & AT & 76.8 & 64.5 & 83.4 & 15.5 & 13.1 & 53.5 & 11.8 & 39.3 & \bf22.7 & 39.3 & 42.0 \\
 \cmidrule{2-13}
\multirow{2}*{SwinL} & Pre-train & {\color{red}\bf86.3} & {\color{red}\bf77.0} & {\color{red}\bf89.6} & {\color{red}\bf31.6} & {\color{red}\bf61.0} & \bf63.6 & {\color{red}\bf26.4} & \bf61.3 & 23.4 & \bf48.8 & \bf56.9 \\
 & AT & 78.7 & 66.9 & 84.9 & 18.2 & 18.1 & 57.3 & 11.6 & 43.4 & \bf25.2 & 42.9 & 44.7 \\
\botrule
\end{tabular*}
\end{minipage}
\end{center}
\end{table*}

\section{Evaluation Results}\label{sec4}

\begin{figure*}[t]
  \centering
\includegraphics[width=0.98\linewidth]{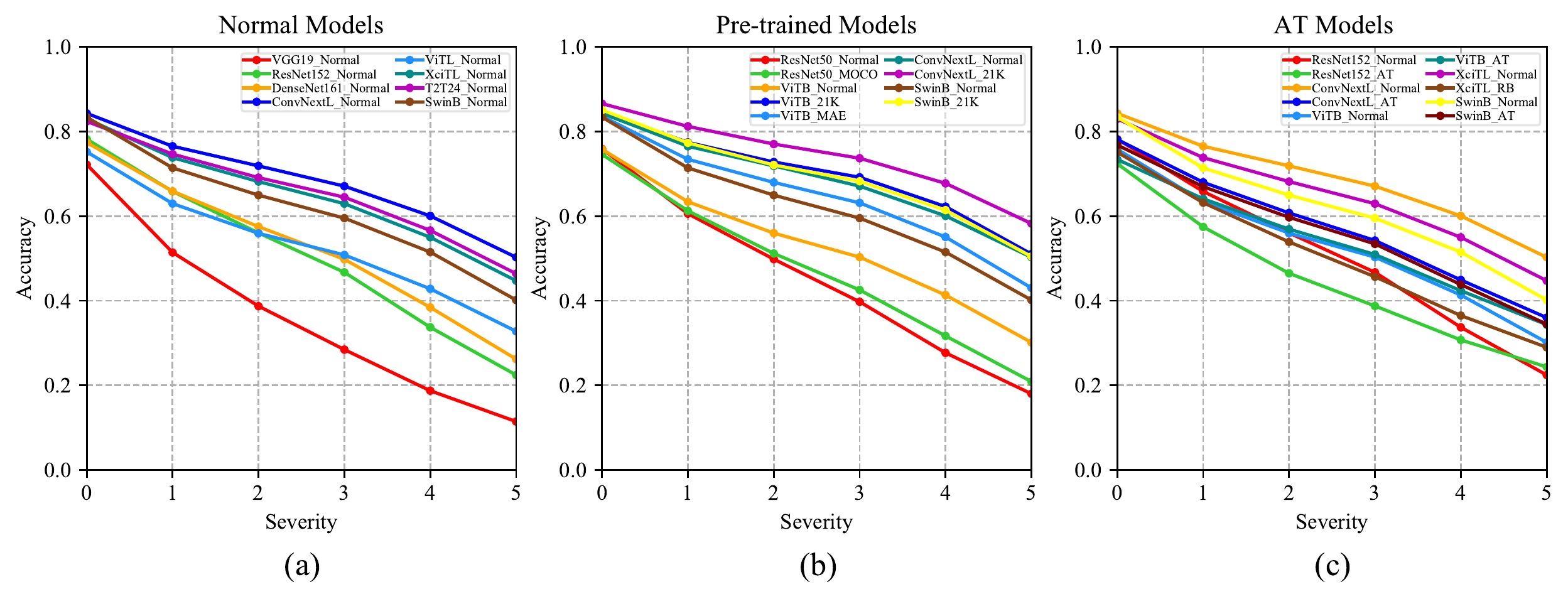}
  \caption{Robustness curves of classification accuracy vs. severity on ImageNet-C. \textbf{(a)} Robustness curves of normally trained models with different architectures, including VGG19, ResNet152, DenseNet161, ConvNextL, ViTL, XciTL, T2T24, and SwinB. \textbf{(b)} Robustness curves of pre-trained models, including 21K-pre-training, MOCOv3, and MAE. \textbf{(c)} Robustness curves of adversarially trained models compared with normally trained models. The model performance is highly consistent under different severities.
  }
  \label{fig:rb-inc}
\end{figure*}

We first present the evaluation of natural robustness in Sec.~\ref{sec:4.1}. Then, the adversarial robustness under white-box and black-box attacks is shown in Sec.~\ref{sec:4.2} and Sec.~\ref{sec:4.3}, respectively. Finally, an interpretability analysis of adversarial training in the frequency domain is provided in Sec.~\ref{sec:4.4}.

\subsection{Natural Robustness Evaluation}\label{sec:4.1}
In this section, we evaluate the natural robustness of the 55 models introduced in Sec.~\ref{sec:3-2-1} under 3 IID and 7 OOD datasets. We show the results of CNNs and Transformers in Table~\ref{tab:ood-cnn} and Table~\ref{tab:ood-transformer}, respectively. Note that all the evaluation results are the Top-1 accuracies, except for ImageNet-C, which uses $\mathrm{mCE}$ as the evaluation metric (lower is better). For consistency, we use $1-\mathrm{mCE}$ as the evaluation metric for ImageNet-C. Besides, on ImageNet-C, we further show the robustness curve of classification accuracy vs. corruption severity in Fig.~\ref{fig:rb-inc}. Detailed results of ImageNet-C robustness curves of all models will be shown in Appendix B.
Based on the experimental results, we have the following observations.

\textbf{Model architecture.} We first compare the natural robustness of different architectures. It is obvious that the natural robustness of Transformers is better than that of most CNNs, except for ConvNexts. ConvNextL based on pre-training achieves 57.4\% natural robustness, which exactly matches the performance of the best Transformer model ViTL. Besides, the normally trained ConvNextL achieves 51.0\% natural robustness, which is better than all Transformers based on normal training. The results demonstrate that CNNs can achieve comparable or even better natural robustness compared with Transformers, which is different from the conclusions of previous works \citep{bai2021transformers,shao2021adversarial,paul2022vision}. Therefore, we think that the key factors of modern architectures, including patchified input images, enlarged kernel size, and reduced activation and normalization layers, are essential to natural robustness rather than the self-attention mechanism \citep{wang2022can}. 
Moreover, a larger model usually leads to better natural robustness within the same architecture family. For example, based on normal training, the natural robustness increases from 34.1\% of ResNet50 to 36.8\% of ResNet101, and finally to 38.0\% of ResNet-152. However, this improvement is not very significant. And for some other architectures, a larger model does not necessarily improve robustness (e.g., 35.8\% of ViTB and 35.4\% of ViTL). From Fig.~\ref{fig:rb-inc}(a), it can be observed that the corruption robustness of typical models is consistent across different severities.

\textbf{Pre-training on ImageNet-21K.} From the results in Table~\ref{tab:ood-cnn} and Table~\ref{tab:ood-transformer}, we can see that pre-training on ImageNet-21K significantly improves natural robustness of CNNs and Transformers. For example, pre-training improves the performance of ViTL from 35.4\% to 57.4\%, exhibiting a large margin. We think using more training data prevents models from overfitting to a certain data distribution, and promotes the natural robustness.

\textbf{Self-supervised learning (SSL).} MOCOv3 achieves slightly better performance than normal training as shown in Table~\ref{tab:ood-cnn}, and MAE improves the natural robustness significantly based on ViT, as shown in Table~\ref{tab:ood-transformer}. It proves that SSL enables models to learn informative representations, which contain fewer spurious cues corresponding to class labels. Thus SSL has better natural robustness. However, the natural robustness of SSL is inferior to pre-training on ImageNet-21K, demonstrating the advantage of using more data. A promising method is performing SSL on larger-scale datasets to further improve natural robustness. 

\begin{figure*}[t]
  \centering
  \includegraphics[width=0.98\linewidth]{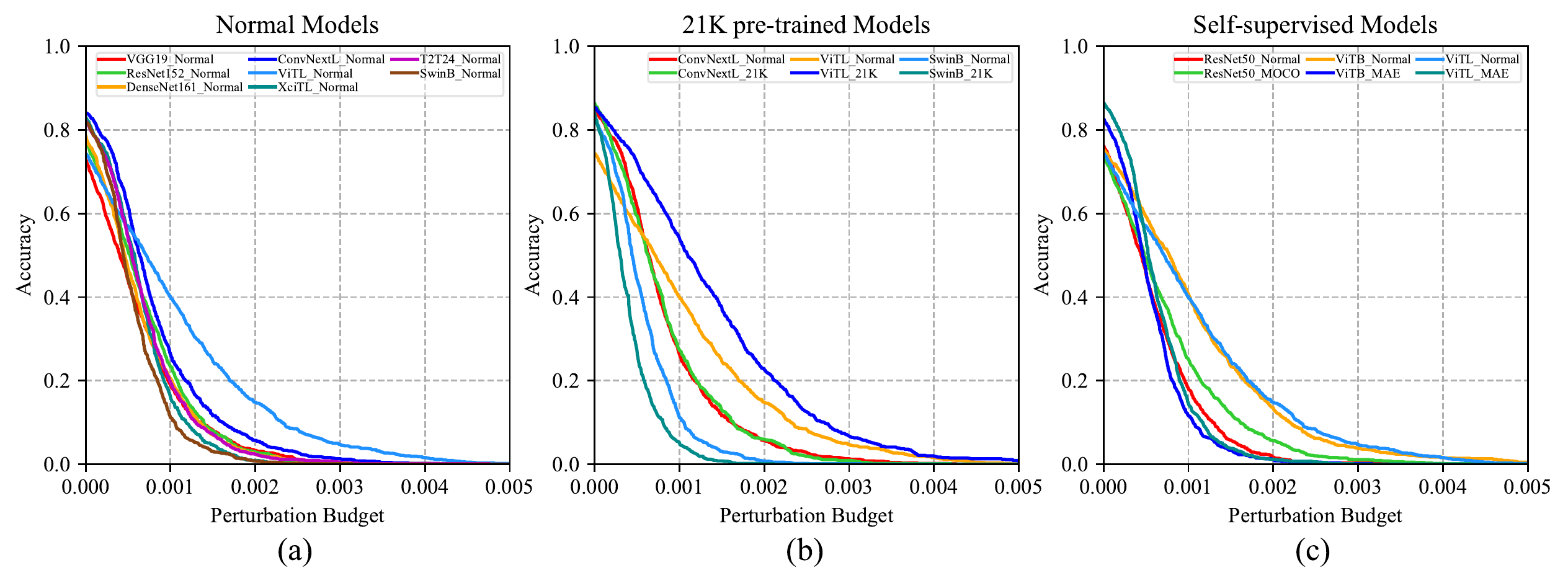}
  \caption{Robustness curves of classification accuracy vs. perturbation budget under AutoAttack. \textbf{(a)} Robustness curves of normally trained models with different architectures, including VGG19, ResNet152, DenseNet161, ConvNextL, ViTL, XciTL, T2T24, and SwinB. \textbf{(b)} Robustness curves of pre-trained models on ImageNet-21K. \textbf{(c)} Robustness curves of self-supervised pre-trained models, including MOCO and MAE. ViT exhibits better adversarial robustness than others.}
  \label{fig:rb-nonadv}
\end{figure*}

\textbf{Adversarial training (AT).} Despite the effectiveness in adversarial robustness, we observe that almost all AT models have significant performance drops in natural robustness, except for ViT. 
Generally speaking, this is because there is a huge shift between adversarial noise and natural noise. For ViT, we conjecture that adversarial examples serve as a form of data augmentation, and compensate for the weak inductive bias of ViT, leading to the improved performance. However, our finding is somewhat contradictory to previous ones \citep{tsipras2018robustness,zhang2019interpreting,ilyas2019adversarial} which deem that AT learns robust and shape-biased features invariant to spurious changes, such as texture and background changes. Thus it can be expected that AT models have better robustness to natural distribution shifts based on previous findings. In fact, AT models lead to better performance on some OOD datasets with different image styles (e.g., ImageNet-R, Stylized-ImageNet, and ImageNet-Sketch), corroborating the previous arguments that AT models learn a more shape-biased representation. 
However, this representation does not generalize well to other real-world distribution shifts, such as viewpoint changes, uncommon objects, etc. Therefore, it still remains a challenge to learn generalizable representations that are not only biased towards shape but also invariant to changes of real-world objects.

\textbf{Comparison of different OOD datasets.}
We find that the performance on several OOD datasets (e.g., ObjectNet, ImageNet-C) are highly consistent with the clean classification accuracy, which is reasonable due to the consistent performance drops. 
Given two models, the probability that the model with lower accuracy
classifies a sample correctly while the other one with higher accuracy classifies it incorrectly is small, if the distribution shift is relatively small \citep{mania2020classifier}.
Besides, ObjectNet, ImageNet-V, and Stylized-ImageNet are harder datasets than the others since the best models still have less than 30\% accuracies on them. It demonstrates that the models are more vulnerable to viewpoint changes and style transfer.

\subsection{White-box Adversarial Robustness Evaluation }\label{sec:4.2}
In this section, we evaluate the adversarial robustness of all models under white-box attacks given the $\ell_\infty$-norm bounded perturbations. The experiments on the $\ell_2$-norm bounded attacks are provided in Appendix D.
The experiments are conducted on 1,000 randomly sampled images from the ImageNet validation set due to the tremendous computation cost of running all data.

\subsubsection{Results on Normally Trained Models}
We first evaluate the adversarial robustness of normally trained models and pre-trained models, while the results of adversarially trained models are presented in Sec.~\ref{sec:4.2.2}. We adopt AutoAttack as the basic attack method for evaluation, which is a powerful white-box attack and widely used in benchmarking adversarial robustness. 
We provide an extensive evaluation on typical model architectures and two pre-training paradigms, including pre-training on ImageNet-21K and self-supervised learning. The detailed robustness curves are shown in Fig.~\ref{fig:rb-nonadv}. We also provide the full results of all models in Appendix C. 

From Fig.~\ref{fig:rb-nonadv}(a), some models with higher clean accuracy, such as SwinB, XciTL, and T2T24, tend to drop more rapidly as the perturbation budget increases. This suggests that the decision boundaries of these models are relatively close to the data points. Therefore, the boundary error introduced in \cite{zhang2019theoretically} will increase more rapidly. Besides, ViT is the most resistant architecture to large perturbations. This is due to the fact that ViT has a smoother loss landscape under input perturbations as discussed in \cite{paul2022vision}. 
The effectiveness of pre-trained models on ImageNet-21K for adversarial robustness is inconsistent, e.g., ViT has better robustness but Swin has worse robustness with pre-training, as shown in Fig.~\ref{fig:rb-nonadv}(b).  This could be due to the fact that ViT has a weaker inductive bias, so the pre-trained models on large-scale datasets learn fewer spurious features, which benefit the adversarial robustness.
For self-supervised learning methods, MAE has a negative impact on the adversarial robustness especially under large perturbation budgets, but MOCOv3 improves adversarial robustness compared with the normally trained models, as shown in Fig.~\ref{fig:rb-nonadv}(c). We believe it is caused by the different learning methods (i.e., contrastive learning of MOCOv3 is more resistant to image noises).

\begin{figure}[t]
  \centering
  \includegraphics[width=0.9\linewidth]{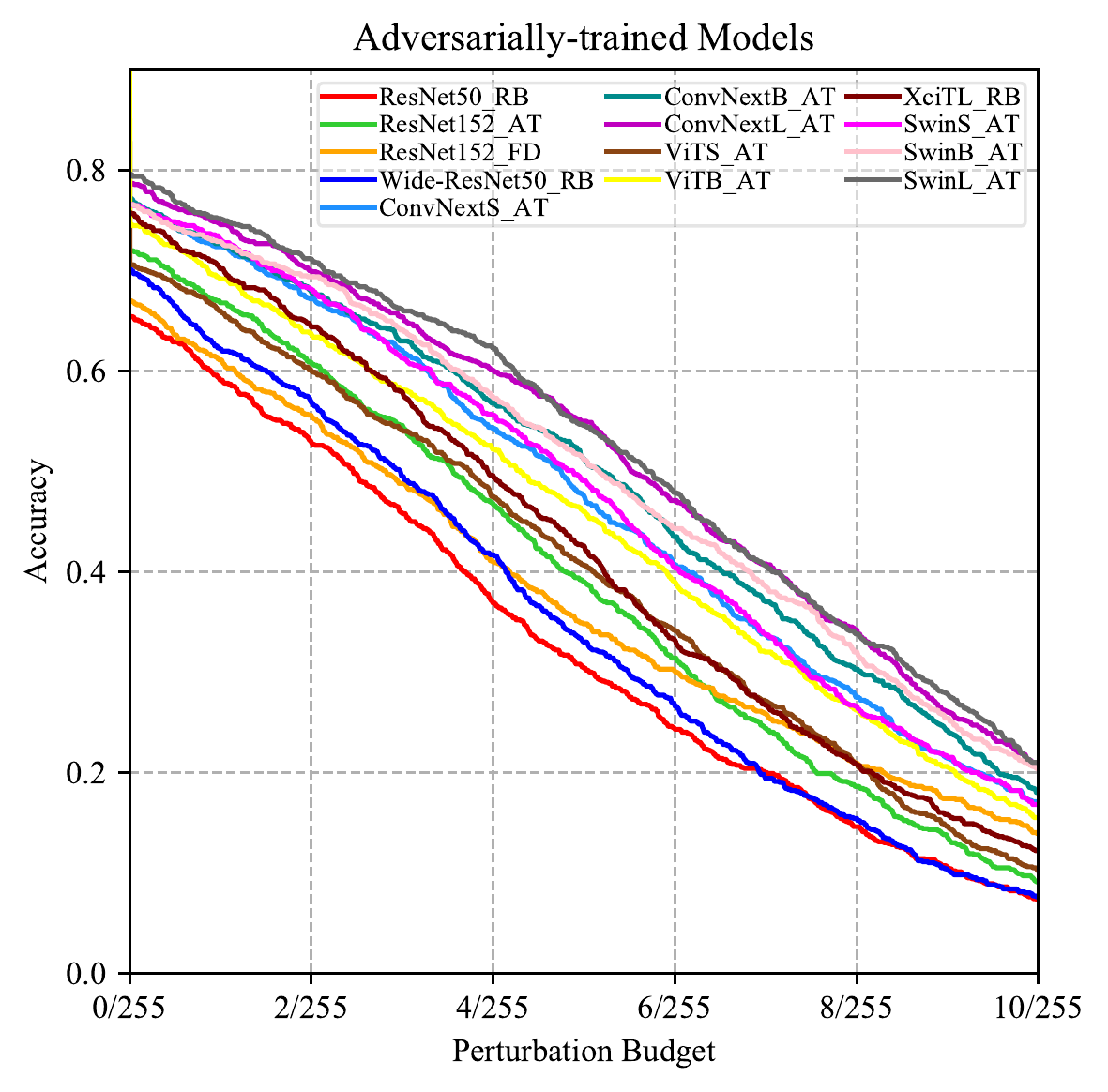}
  \caption{Robustness curves of adversarially trained models with different network architectures, including ResNets, ConvNexts, ViTs, XciTs, and Swins under AutoAttack. ConvNextL performs best in CNNs, and SwinL performs best in Transformers.}
  \label{fig:rb-adv-arch-ens}
\end{figure}

\begin{table*}[t]\footnotesize
\caption{White-box robustness (\%) with different architectures. All the results are obtained by adversarial attacks with  $\epsilon=4/255$. ConvNextL performs best in CNNs, and SwinL performs best in Transformers.}
\label{tab:rb-adv-arch}
\begin{center}
\begin{minipage}{0.75\textwidth}
\begin{tabular*}{\textwidth}{@{\extracolsep{\fill}}llcccc@{\extracolsep{\fill}}}
\toprule%
Architecture & Method & Clean Acc & FGSM & PGD100 & AutoAttack \\
\midrule
\multirow{3}*{ResNet50} & AT & 67.0 & 44.5 & 38.7 & 34.1 \\
  & RB & 65.5 & 46.7 & 42.1 & 36.9 \\
 & RL & 63.4 & 42.7 & 34.3 & 30.9 \\
 \cmidrule{2-6}
ResNet101 & AT & 71.0 & 51.3 & 46.5 & 42.2 \\
\cmidrule{2-6}
\multirow{2}*{ResNet152} & AT & 72.4 & 54.6 & 49.6 & 46.7 \\
 & FD & 67.0 & 51.6 & 48.8 & 41.0 \\
\cmidrule{2-6}
\multirow{2}*{Wide-ResNet50} & AT & 70.5 & 51.8 & 44.6 & 39.3 \\
 & RB & 70.2 & 51.1 & 45.2 & 41.7 \\
 \midrule
 ConvNextS&AT&77.3&60.3&56.9&54.3\\
 \cmidrule{2-6}
  ConvNextB&AT&77.2&62.2&59.0&56.8\\
 \cmidrule{2-6}  ConvNextL&AT&\bf78.8&\bf63.9&\bf61.7&\bf60.1\\
\midrule
ViTS & AT & 70.7 & 51.3 & 47.5 & 43.7 \\
 \cmidrule{2-6}
ViTB & AT & 74.7 & 55.9 & 52.2 & 49.7 \\
\midrule
XciTS & RB & 74.5 & 51.6 & 46.0 & 43.1 \\
 \cmidrule{2-6}
XciTM & RB & 75.3 & 54.8 & 50.2 & 47.0 \\
 \cmidrule{2-6}
XciTL & RB & 75.8 & 55.7 & 57.1 & 49.6 \\
\midrule
 SwinS & AT & 76.6 & 61.5 & 58.4 & 55.6 \\
 \cmidrule{2-6}
SwinB & AT & 76.6 & 63.2 & 60.2 & 57.3 \\
 \cmidrule{2-6}
SwinL & AT & \bf79.7 & \bf65.9 & \bf63.9 & \bf62.3 \\
\botrule
\end{tabular*}
\end{minipage}
\end{center}
\end{table*}

However, the adversarial robustness of normally trained models, including pre-trained ones, is relatively worse than adversarially trained models. Therefore, the conclusions are only valid for normal training.

\subsubsection{Results on Adversarially Trained Models}\label{sec:4.2.2}

We then evaluate the adversarial robustness of AT models. AT is such an effective method to improve robustness that most of the state-of-the-art robust models under white-box attacks are obtained by AT. It is shown that AT can be used to explore the upper limit of model robustness. 

To comprehensively show the performance of AT models, we also exhibit the robustness curves of several models in Fig.~\ref{fig:rb-adv-arch-ens}. More detailed results of all models are provided in Appendix C. Besides, we also show the robustness under various white-box attacks with $\epsilon=4/255$ in Table~\ref{tab:rb-adv-arch}. 
Based on the experimental results, we have the following observations.

\begin{table*}[t]\footnotesize
\caption{Adversarial robustness (\%) of different training tricks in AT. Most training tricks, including RandAugment (RA), Mixup, label smoothing (LS), weight decay (WD) and EMA, are studied on SwinS. The pre-training methods, including 21K-pre-training and SimMIM, are studied on SwinB. The pre-training method CLIP is studied on ViTB. A combination of RA, Mixup, LS (0.1), and EMA performs best in SwinS. The models pre-trained with ImageNet-21K and SimMIM have similar performance as the model trained from scratch. However, the model pre-trained with CLIP has a performance drop compared with the model trained from scratch.}
\label{tab:rb-adv-strategy}
\begin{center}
\begin{minipage}{0.98\textwidth}
\begin{tabular*}{\textwidth}{@{\extracolsep{\fill}}ccccccccccc@{\extracolsep{\fill}}}
\toprule%
\multirow{3}*{Model} & \multicolumn{6}{c}{Strategy} & \multirow{3}*{Clean Acc} & \multirow{3}*{FGSM} & \multirow{3}*{PGD100} & \multirow{3}*{AutoAttack} \\
\cmidrule{2-7}
& RA & Mixup & LS & WD & EMA & Pre-training  \\
\midrule
\multirow{8}*{SwinS}& \xmark & \xmark & 0.0 & 0.0 & \xmark &\xmark &73.9 & 53.8 &50.3 &48.7\\
&\cmark &\xmark &0.0 &0.0 &\xmark &\xmark & 74.5 &55.7 &52.0&49.7\\
&\cmark &\cmark &0.0 &0.0 &\xmark &\xmark & 76.4 &60.4 &56.8&54.9\\
&\cmark &\cmark &0.1 &0.0 &\xmark & \xmark& 76.9 &61.4 &57.9&55.4\\
&\cmark &\cmark &0.1 &0.01 &\xmark & \xmark& \bf77.1 & 60.5 &57.7&54.7\\
&\cmark &\cmark &0.1 &0.05 &\xmark & \xmark& 76.7 & 60.8 &57.7&54.4\\
&\cmark &\cmark &0.1 &0.1 &\xmark & \xmark& 76.7 &58.3 &56.3 &53.7\\
&\cmark &\cmark &0.1 &0.0 &\cmark & \xmark & 76.6 & \bf61.5 & \bf58.4& \bf55.6 \\

\midrule
\multirow{3}*{SwinB} &\cmark &\cmark &0.1 &0.0 &\cmark & \xmark & 76.6 &73.4 &\bf60.2 &57.3 \\
&\cmark &\cmark &0.1 &0.0 &\cmark & 21K & \bf77.5 &\bf74.8 &59.2&\bf57.4\\
&\cmark &\cmark &0.1 &0.0 &\cmark & SimMIM & 77.2 &73.6 &59.1&55.9\\
\midrule
\multirow{2}*{ViTB} &\cmark &\cmark &0.1 &0.0 &\cmark & \xmark & \bf74.7 & \bf69.2 & \bf52.2 & \bf49.7 \\
&\cmark &\cmark &0.1 &0.0 &\cmark & CLIP & 69.0 & 63.3 & 46.3 & 42.8\\
\botrule
\end{tabular*}
\end{minipage}
\end{center}
\end{table*}

From Fig.~\ref{fig:rb-adv-arch-ens}, it can be seen that the robustness trends of different architectures under different perturbation budgets are  consistent. First, the most robust Transformer model is SwinL, which achieves 62.3\% robust accuracy under AutoAttack with $\epsilon=4/255$. The most robust CNN model is ConvNextL with 60.1\%  robust accuracy under the same setting. Both results are clearly state-of-the-art compared with XciTL in RobustBench (with 49.6\% robust accuracy under AutoAttack). In contrast to previous debates on whether Transformers have better adversarial robustness than CNNs \citep{tang2021robustart,aldahdooh2021reveal,shao2021adversarial}, we have found that CNNs can achieve competitive (albeit slightly worse) adversarial robustness as Transformers. Therefore, we believe that modern architectures, including patchified input images, enlarged kernel size, and reduced activation and normalization layers, play the most important role in adversarial robustness. Second, there is a trade-off in the perturbation budgets. ResNet152-FD was trained on adversarial examples with $\epsilon=16/255$, resulting in lower robust accuracy (41.0\%) under AutoAttack with $\epsilon=4/255$ than ResNet152-AT (46.7\%). However, the robustness of ResNet-FD under a larger budget (e.g., $\epsilon=8/255$) is better than that of ResNet152-AT. This suggests that we still need robustness curves for  a comprehensive evaluation of robustness, rather than a single robust accuracy under a fixed perturbation budget. Third, adversarial robustness is improved with a larger model size. A larger model size means a larger model capacity, which has a positive effect on adversarial robustness. Moreover, the depth design of large models, such as SwinL (with a depth of 2-2-18-2) and ConvNextL (with a depth of 3-3-27-3) is consistent with the conclusion in \cite{huang2021exploring} that reducing capacity in the last stage benefits the adversarial robustness.

\begin{figure*}[t]
  \centering
  \includegraphics[width=0.8\linewidth]{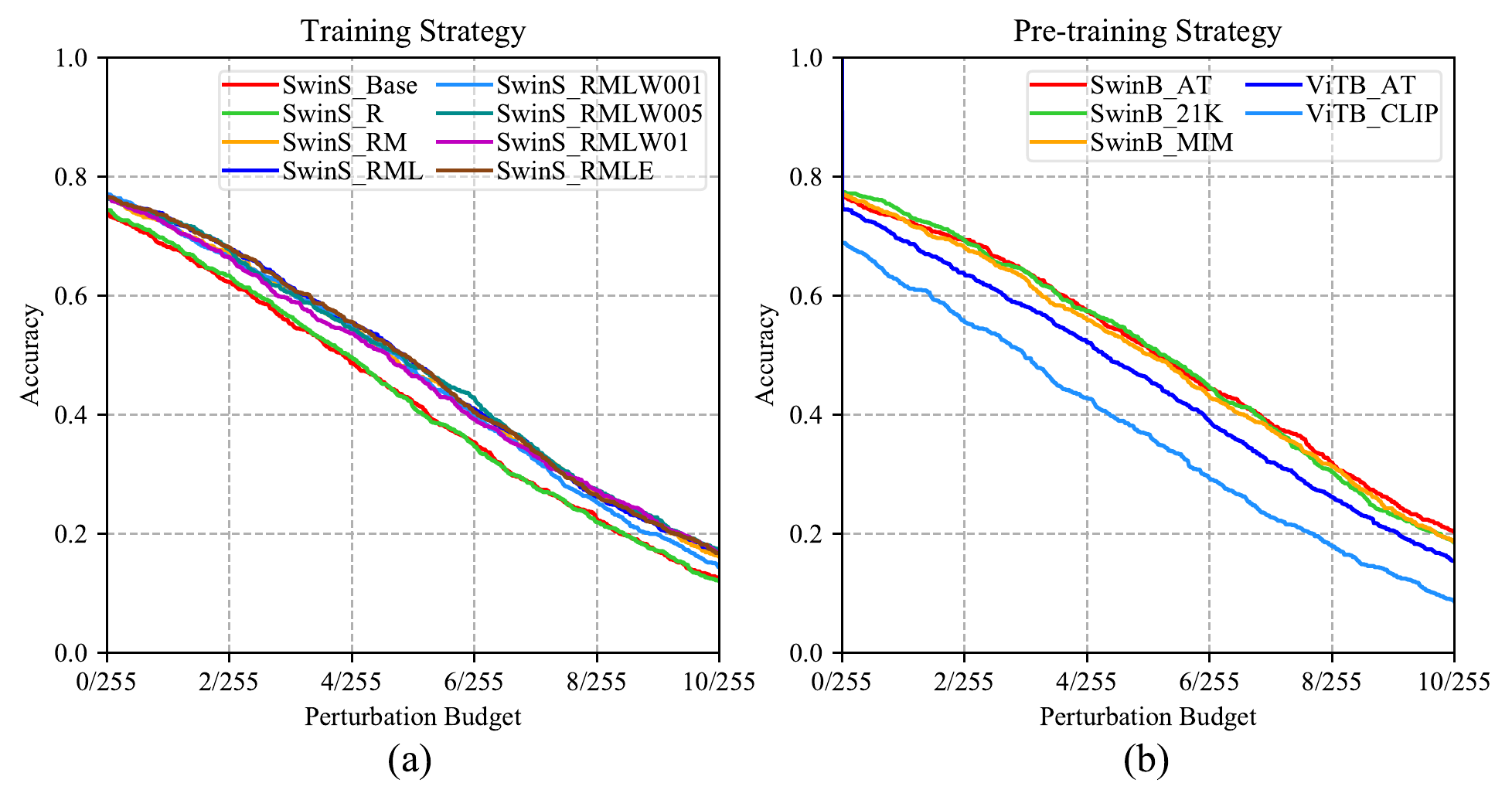}
  \caption{Robustness curves of adversarially trained models with different training tricks. \textbf{(a)} The curves of some training tricks, including weight decay (W), label smoothing (L), RandAugment (R), Mixup (M), and EMA (E). SwinS\_Base is trained without any of the tricks. The rest of the models are trained with the tricks indicated by the capital initial letter and appended to the name of SwinS (e.g., SwinS\_RMLW001 is trained with RandAugment, Mixup, label smoothing and weight decay 0.01). (b) The curves of pre-training methods including 21K-pre-training and SimMIM of SwinB, and CLIP of ViTB. The robustness curves are nearly parallel to each other, and the robustness is consistent with the results in Table~\ref{tab:rb-adv-strategy}.}
  \label{fig:rb-adv-strategy}
\end{figure*}

\subsubsection{Ablation Study on Training Tricks in AT}
In this section, we examine the training tricks in large-scale AT. Since some of the recently proposed tricks are not suitable for the previously proposed networks, such as ResNets, we choose SwinS, SwinB and ViTB (intended for CLIP) as the backbone networks. The detailed results are shown in Table~\ref{tab:rb-adv-strategy}. To comprehensively show the robustness under all perturbation budgets, we also show the robustness curves of these AT models in Fig.~\ref{fig:rb-adv-strategy}. In the ablation study, we adopt the control variate method by changing one type of tricks each time, while leaving the other settings unchanged.

We first train a baseline model with none of the strategies added, which obtains 48.7\% robust accuracy under AutoAttack. For data augmentations, the adversarial robustness improves by 1.0\% with RandAugment and by another 5.2\% with Mixup. AT easily overfits to certain attack patterns \citep{pang2020bag}, so data augmentations can effectively reduce overfitting and improve the adversarial robustness. For label smoothing, the adversarial robustness improves by 0.5\% with appropriate label smoothing. Compared with hard labels, soft labels prevent models from making over-confident decisions, which improves robustness in the presence of noisy inputs and labels \citep{stutz2020confidence,dong2022exploring}. For weight decay, we find that a larger weight decay degrades the robustness, which is different from previous study \citep{pang2020bag}. We think that adequate data augmentation and advanced architectural components are sufficient to avoid overfitting, such that a large weight decay would affect the learning ability of the model. 
For EMA, the robust accuracy improves by 0.2\% with EMA. Using EMA can smoothen the update of model parameters and avoid fluctuations caused by mini-batch training methods. This can make the model more robust on test data.

\begin{figure}[!t]
  \centering
  \includegraphics[width=0.98\linewidth]{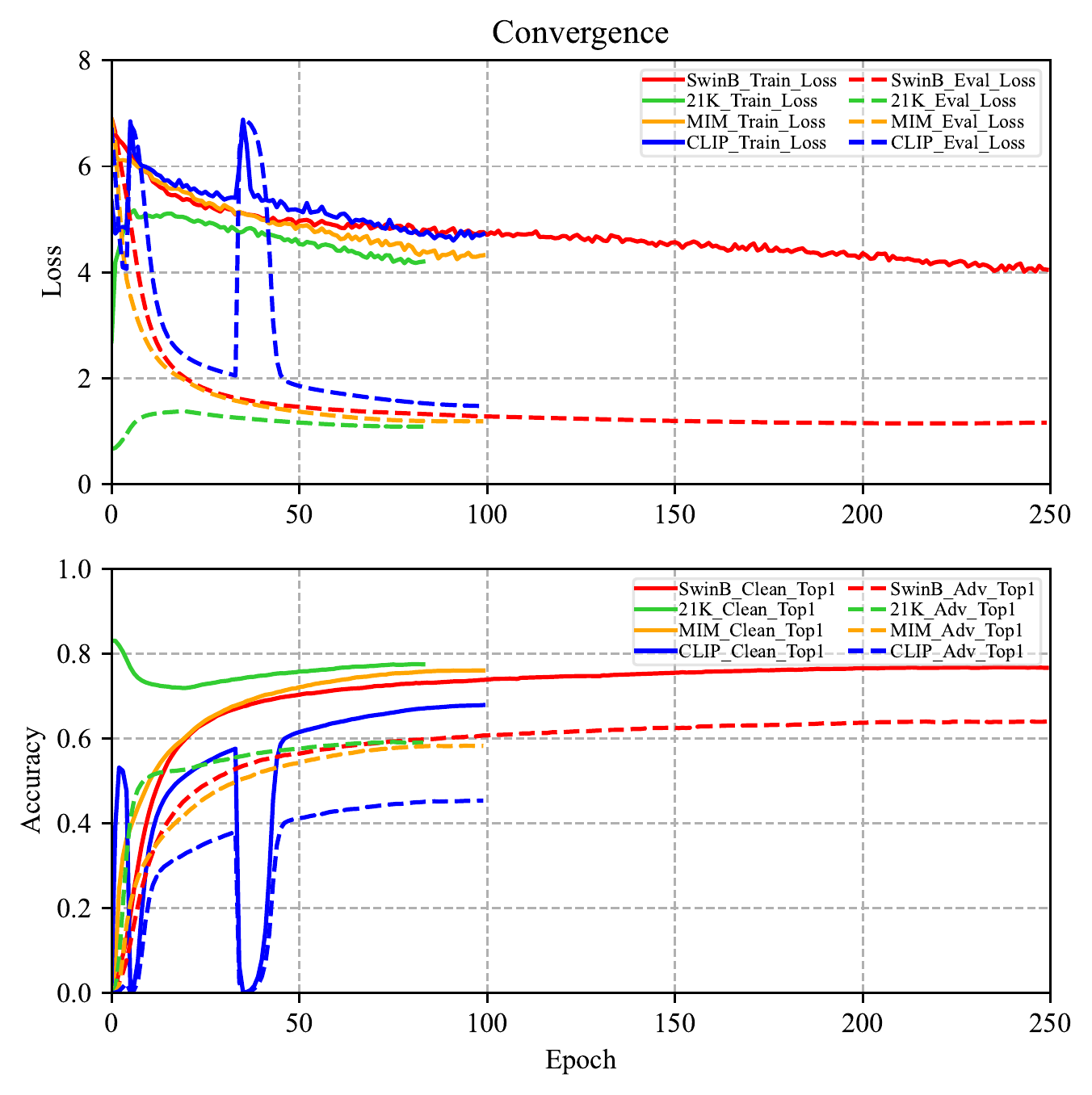}
  \caption{Training curves of vanilla AT, 21K-pre-training and SimMIM pre-training on SwinB and CLIP pre-training on ViTB. Pre-training with ImageNet 21K dataset accelerates AT. The model pre-trained with CLIP is reconstructed during the fine-tuning process, leading to the catastrophic performance drops.}
  \label{fig:rb-time}
\end{figure}

\begin{figure*}[!t]
  \centering
  \includegraphics[width=\linewidth]{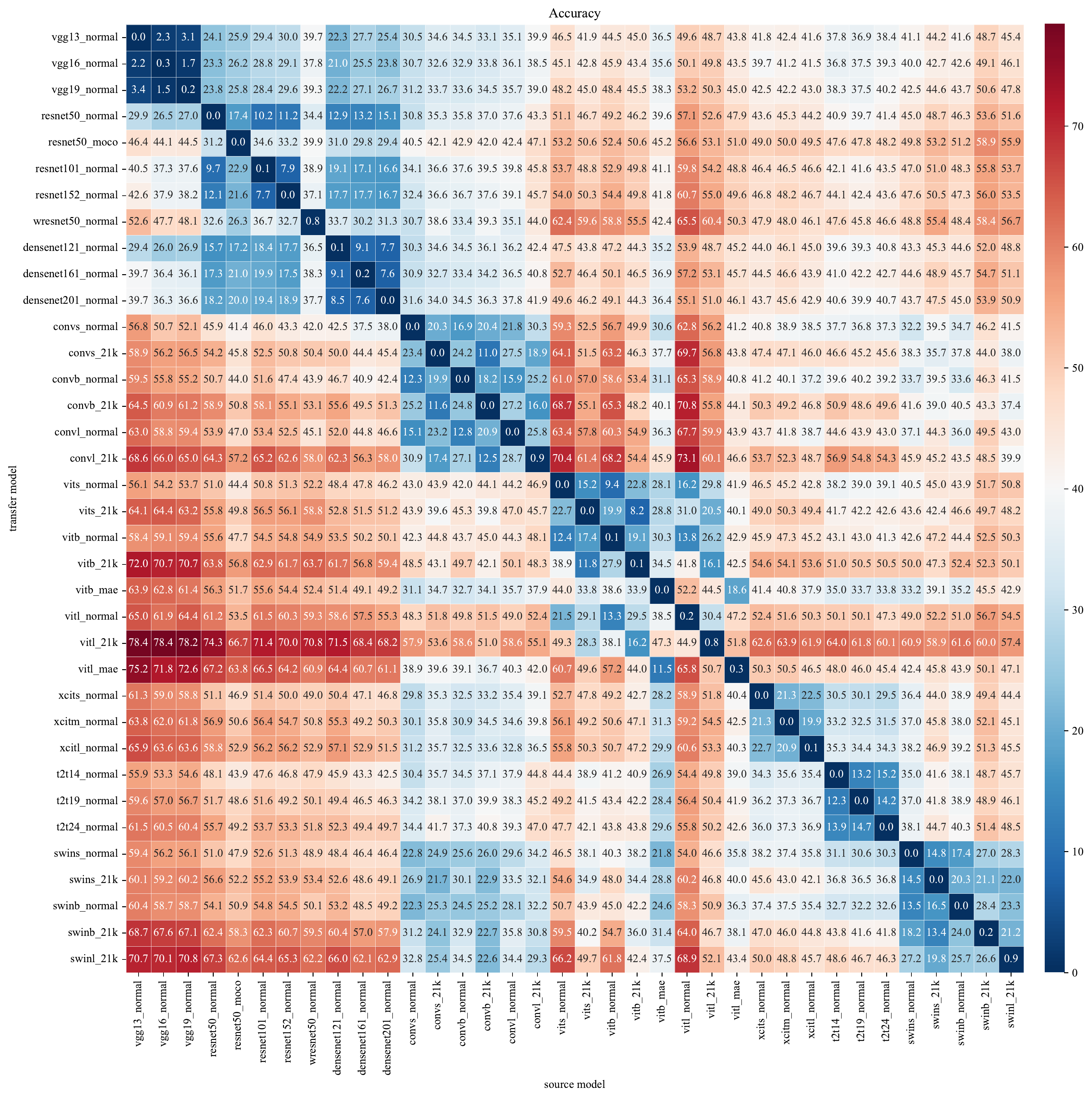}
  \caption{Black-box transferability across normally trained, 21K pre-trained and self-supervised pre-trained models. ViTs are generally more robust than other models. ConvNexts act as the best surrogate models. Transformers generally have better black-box robustness, compared with CNNs. Transferability of adversarial examples between CNNs and Transformers is relatively lower than that within similar architectures.}
  \label{fig:transfer-normal-vmi}
\end{figure*}

We also conduct experiments on some pre-training paradigms including ImageNet 21K pre-training, SimMIM, and CLIP. We adopt the pre-trained weights as the initialization of AT. The main advantage of (self-)supervised pre-training mainly is that it can reduce the convergence time during model training. Therefore, we additionally provide the detailed convergence process in Fig.~\ref{fig:rb-time}, in which the models are tested by the PGD-10 attack with $\epsilon=4/255$ and step size 1/255. For 21K-pre-training, it can be seen that the fine-tuning process is obviously accelerated and the fine-tuned model in the 100-th epoch has competitive robustness with the model trained from scratch. It indicates that the 21K-pre-training learns useful representations that can be beneficial to the fine-tuning process. However, the accuracy on clean samples initially decreases and then increases to a lower level. This means that the fine-tuning process still suffers from catastrophic forgetting. For SimMIM, it can be seen that the robust accuracy curves of the fine-tuned models from SimMIM are close to the curves of the models trained from scratch. It indicates that the representative features learned in SimMIM are not beneficial to the downstream AT. For CLIP, we have witnessed a drop in performance with the image encoder as initialization. Besides, by analyzing the loss and accuracy during model training, it seems that the CLIP model is reconstructed during the training process, which makes the final results do not adequately fit to adversarial examples. This is because the CLIP models learn from the textual information. Once the image encoder is separated from the original architecture, it needs to be reconstructed to match adversarial training.

\subsection{Black-box Adversarial Robustness Evaluation}\label{sec:4.3}
In this section, we evaluate the adversarial robustness of normally trained and adversarially trained models under the black-box attack setting.

\subsubsection{Results on Normally Trained Models}
Here, we use transferability heatmaps to illustrate the robustness of models against black-box attacks instead of the robustness curves.
We use VMI-FGSM as the black-box attack method. The attack transferability heatmap of the normally trained models is shown in Fig.~\ref{fig:transfer-normal-vmi}, in which the source models are the surrogate models of an attack method, and the target models are those models to be tested with the adversarial examples generated by attacking the source models with VMI-FGSM. Additional results of different adversarial attacks are shown in Appendix~E.

From the results, we have the following findings. First, ViTs are generally more robust than other models under black-box attacks, which is consistent with the results of white-box robustness, i.e., among normally trained models, ViT is more resistant to adversarial perturbations. Second, ConvNext models act as the best surrogate models. This means, the adversarial examples generated for them achieve much higher attack success rates than other models. It indicates that ConvNext contains both the gradient information of the convolution in CNNs and that of the structure in Transformers. Third, similar to white-box robustness, Transformers generally perform much better against transfer-based attacks than CNNs. Fourth, models of one architecture can, to some extent, defend against perturbations generated from models of different architectures. It is due to the significant difference of gradients between architectures.

\begin{figure}[t]
  \centering
  \includegraphics[width=\linewidth]{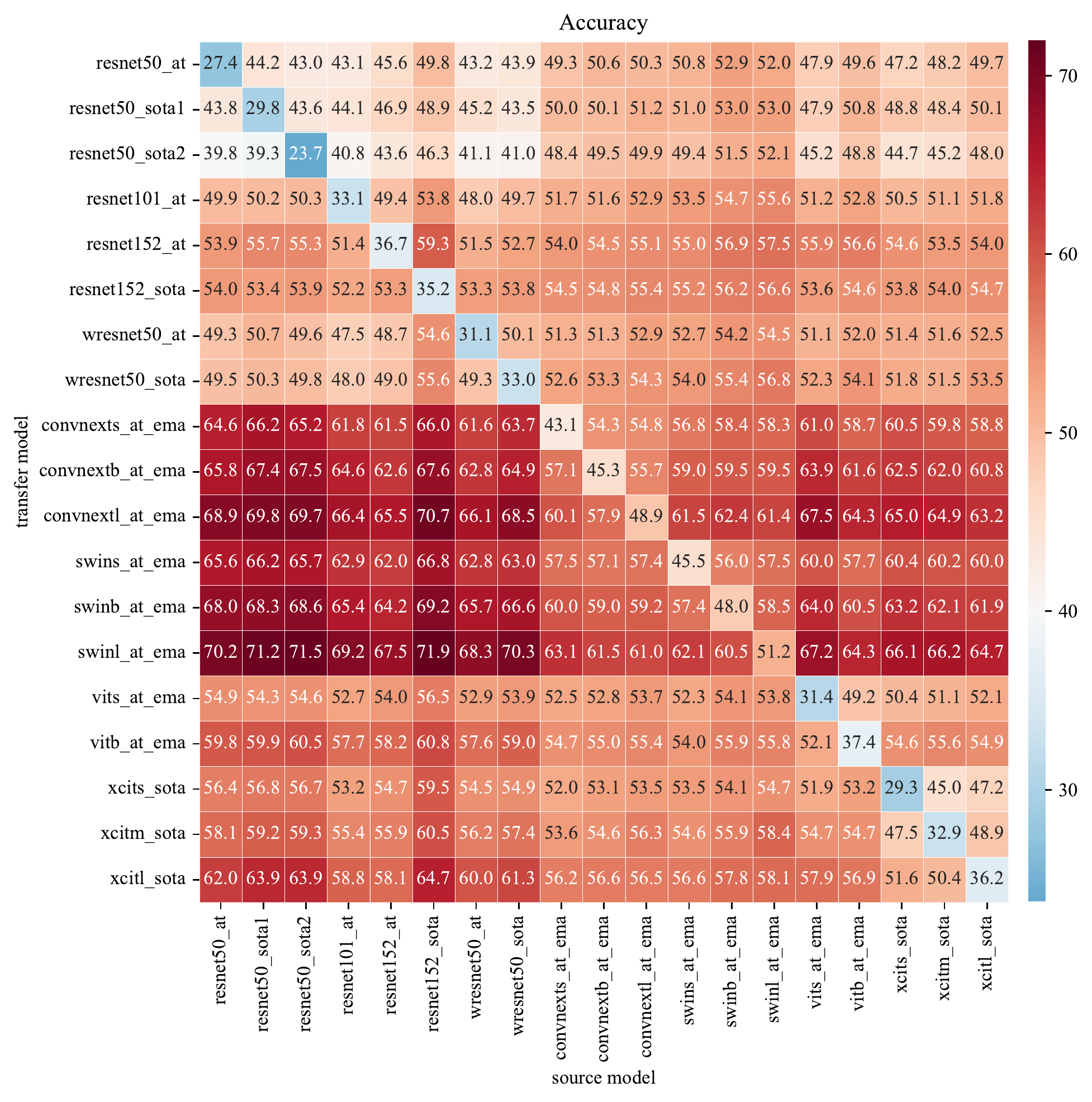}
  \caption{Black-box transferability across adversarially trained models. The models with higher white-box robustness generally have better black-box robustness. }
  \label{fig:transfer-adv-vmi}
\end{figure}

\subsubsection{Results on Adversarially Trained Models}
We then evaluate the robustness of the adversarially trained models under transfer-based black-box attacks.
The detailed results of VMI-FGSM are shown in Fig.~\ref{fig:transfer-adv-vmi}. Additional results of different adversarial attacks are shown in Appendix E.

From the results, we have the following findings. First, the robustness to transfer-based attacks is significantly improved by AT. Models with higher white-box robustness also have a better performance against black-box attacks. Second, ConvNexts have the best robustness under transfer-based attacks among CNNs. Swins have the best robustness under transfer-based attacks among Transformers. It is highly consistent with white-box robustness.

\subsection{Relationship Between Frequency bias and Robustness}
\label{sec:4.4}

\renewcommand{\arraystretch}{1.0}
\begin{table}[t]\footnotesize
\begin{center}
\caption{Frequency bias $f_{bias}$ in Eq.~\eqref{eq:frenq} of normally trained and adversarially trained models, including ResNet, ViT, XciT and Swin. Adversarially trained models have lower frequency bias, compared with the normally trained ones.}
\label{tab:fc-bias}%
\begin{tabular}{@{}llcc@{}}
\toprule
Architecture & Model & Normal & AT \\
\midrule
\multirow{3}*{ResNet} &ResNet50 & 32.1& 22.8   \\
&ResNet101 & 30.0 & 24.9 \\
&ResNet152 & 28.3 & 23.3 \\
\midrule
\multirow{3}*{ConvNext} &ConvNextS & 29.9& 23.9   \\
&ConvNextB & 28.1 & 22.5 \\
&ConvNextL & 26.7 & 21.9 \\
\midrule
\multirow{2}*{ViT} & ViTS & 29.1 & 22.5 \\
& ViTB & 28.5 & 21.3 \\
\midrule
\multirow{3}*{XciT} & XciTS & 28.3 & 23.8 \\
& XciTM & 28.0 & 24.6 \\
& XciTL & 28.6 & 25.7 \\
\midrule
\multirow{3}*{Swin} & SwinS & 30.5 & 21.6 \\
& SwinB & 28.1 & 21.6 \\
& SwinL & 24.9 & 20.8 \\
\botrule
\end{tabular}
\end{center}
\end{table}

\begin{figure*}[t]
  \centering
  \includegraphics[width=0.93\linewidth]{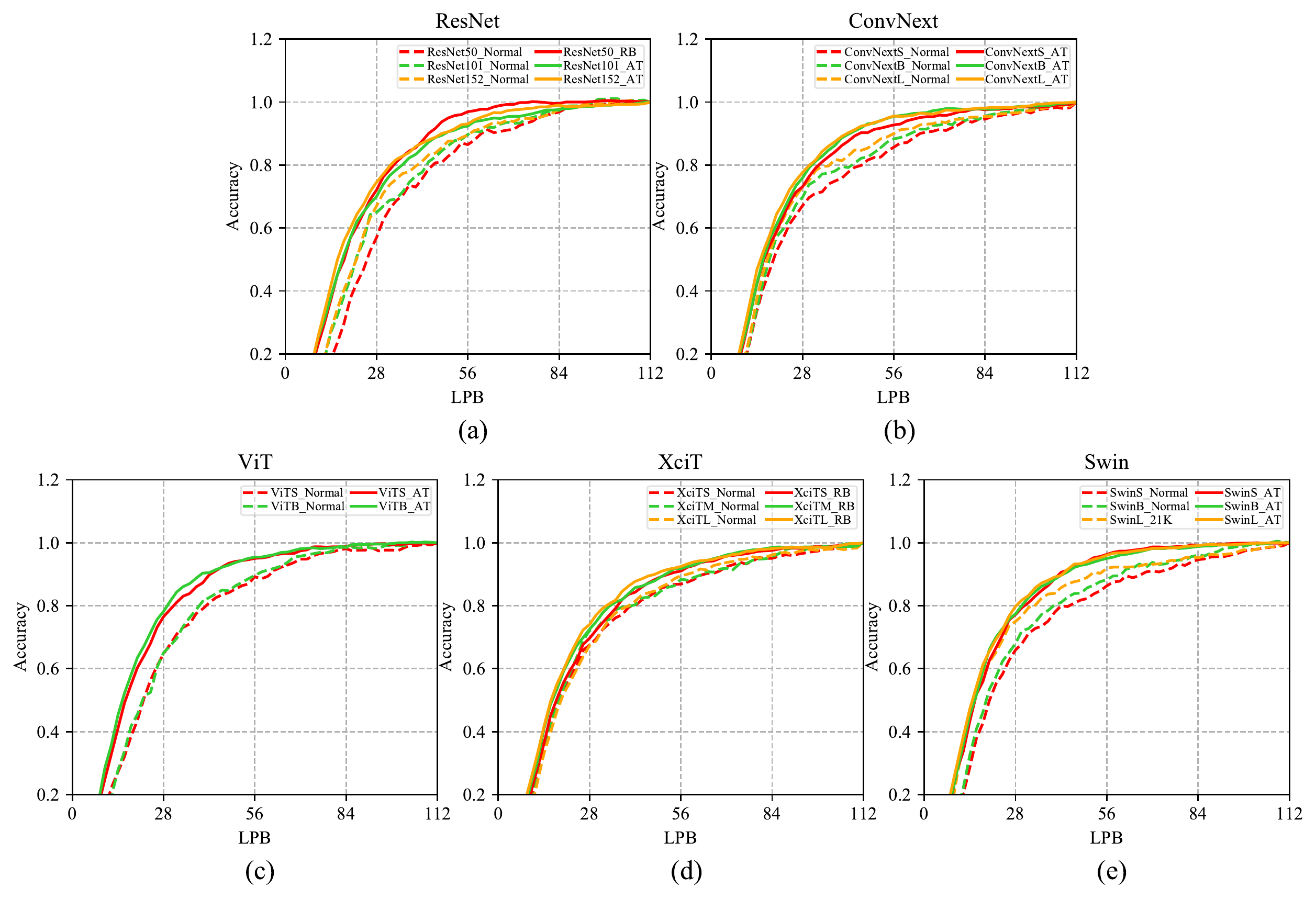}
  \caption{The ACC-LPB curves of models with architectures ResNet, ViT, XciT and Swin. The accuracy of each architecture is normalized to 0-1. The higher robust curve means lower frequency bias of the target model. Adversarially-trained models generally appear above normally-trained ones.}
  \label{fig:fc-curve}
\end{figure*}

In this section, we try to understand the robustness of the models from the perspective of the frequency domain. We use the ACC-LPB curve and the frequency bias metric as introduced in Sec.~\ref{sec:3.2.3}. We mainly compare the differences between normally trained and adversarially trained models to investigate the frequency bias of AT. The ACC-LPB curve is shown in Fig.~\ref{fig:fc-curve}. For the convenience of comparison, we normalize the accuracy to the range of 0-1. We also calculate the frequency bias of these two types of models, which is shown in Table~\ref{tab:fc-bias}.

The higher the LPB, the more high-frequency information is added to the images, increasing the accuracy.  The region with a higher growth rate means that the models pay more attention to the information within that region. Compared with the curves of the normally trained models, the curves of the AT models increase more rapidly within the lower LPB region, and become flatter within the higher LPB region. It means that the AT models pay more attention to low-frequency information. The result is consistent with the frequency bias calculated in Table~\ref{tab:fc-bias} which states that the AT models have lower frequency bias than normal ones. Low-frequency information commonly contains the shapes of objects in an image. It is concluded that AT forces models to learn more shape-biased features. This conclusion is also consistent with the natural robustness in Table~\ref{tab:ood-cnn} and Table~\ref{tab:ood-transformer}. Compared with normally trained modes, the performance of AT models is improved on some specific OOD datasets, such as ImageNet-R, Stylized-ImageNet and ImageNet-Sketch. The images in these datasets have lost much more detailed textures in a high-frequency domain, due to the style transfer. In this paper, we utilize the frequency analysis only as a tool to explain the frequency attention bias of AT. It can be developed as a regularization method to achieve better adversarial robustness in future work.

\section{Conclusions and Discussions}\label{sec5}
In this paper, we established a comprehensive and rigorous benchmark to evaluate the natural and adversarial robustness of image classifiers. We performed large-scale experiments on CNN and Transformer models trained with four paradigms, including normal training, pre-training on large-scale datasets, self-supervised learning, and adversarial training. We evaluated the natural robustness on 10 datasets and also reported the robustness curves under the state-of-the-art adversarial attack of AutoAttack. Besides, we also
conducted ablation studies on the training tricks in large-scale adversarial training and provided an optimized setting. Finally, we provided a frequency perspective to show the frequency bias of adversarially trained models.
Based on the results, we highlight some important findings.

First, there exists a trade-off between adversarial and natural robustness given a model architecture. Typically, adversarial training significantly degrades the natural robustness on most OOD datasets, although the adversarial robustness is improved. The trade-off found by us complements the well-known accuracy-robustness trade-off \citep{zhang2019theoretically}, while contradicting some previous findings \citep{tsipras2018robustness,zhang2019interpreting,ilyas2019adversarial}. As discussed in Sec.~\ref{sec:4.1}, we think that although adversarial training enables the learning of more shape-biased and human-interpretable representations, they do not generalize well to real-world distribution shifts, especially changes in viewpoint and style transfer. Our results suggest that adversarial training is not a universally applicable solution for improving model robustness. It still remains an open problem of how to achieve both natural and adversarial robustness in tandem. 

Second, although Transformers outperform most CNNs in terms of  natural robustness and adversarial robustness, the more advanced ConvNext models achieve comparable robustness to Transformers. Specifically, ConvNext achieves very similar natural robustness and slightly worse adversarial robustness than the best Transformer. In contrast to previous findings that Transformers achieve superior robustness over CNNs \citep{bhojanapalli2021understanding,bai2021transformers,tang2021robustart,aldahdooh2021reveal,shao2021adversarial},
we find that modern architectural designs, including patchified input images, enlarged kernel size, and reduced activation and normalization layers, are essential to robustness, rather than the self-attention mechanism.

Third, pre-training on large-scale datasets (e.g.,  ImageNet-21K) or by self-supervised learning significantly improves natural robustness. More training data can prevent models from overfitting to the data distribution and lead to better generalization performance, while self-supervision can make models learn predictive features with fewer spurious cues, which also exhibit improved natural robustness. Besides, we find that the pre-trained models on larger datasets can serve as better initializations to speed up the fine-tuning process of adversarial training. Based on the normal pre-trained models on ImageNet-21K, we can achieve the state-of-the-art adversarial robustness with a larger model size without incurring huge computational costs. 

Finally, adversarial training generally suffers from the problem of overfitting \citep{rice2020overfitting}. Therefore, some of the training tricks, such as data augmentation, regularization, weight averaging, can improve adversarial robustness by mitigating overfitting in adversarial training. Based on the ablation studies, we provide a recipe for training robust models on ImageNet with appropriately designed tricks.

\section*{Statements and Declarations}
\begin{itemize}
\item Competing interests.

The authors declare that they have no known competing financial interests or personal relationships that could have appeared to influence the work reported in this paper.
\end{itemize}

\renewcommand*{\bibfont}{\small} 
\bibliography{sn-bibliography}

\end{document}


\title[ARES-Bench]{A Comprehensive Study on Robustness of Image Classification Models: Benchmarking and Rethinking}

\maketitle


\begin{appendices}

\section{Experimental settings}\label{app:a}
In this part, we provide the detailed settings for both adversarial attacks and model training in our paper.

For adversarial attacks, we utilize FGSM~\citep{goodfellow2014explaining}, PGD100~\citep{Madry2017Towards}, and AutoAttack~\citep{croce2020reliable} for white-box robustness evaluation; MIM~\citep{dong2018boosting}, DIM~\citep{xie2019improving}, TIM~\citep{dong2019evading}, SI-NI-FGSM~\citep{lin2019nesterov}, and VMI-FGSM~\citep{wang2021enhancing} for black-box robustness evaluation. For $L_\infty$ white-box attacks, the attack epsilon for all white-box attacks is 4/255. For FGSM, the attack step size is 4/255. For PGD100, the attack step is 100, the attack step size is 1/255, and the attack starts with an random noise. For AutoAttack, we adopt the standard version of attack setting. For MIM, the attack step is 20 and the decay factor is 1.0. For $L_\infty$ black-box attacks, the attack epsilon for all black-box attacks is 8/255. For DIM, the attack step is 20 and the decay factor is 1.0. The input images are augmented by random resize and crop with probability 0.7. For TIM, the attack step is 20 and the decay factor is 1.0. The kernel filter is a Gaussian kernel with size 15 and sigma 3. For SI-NI-FGSM, the attack step is 20, decay factor is 1.0, the scale factor is 5. For VMI-FGSM, the attack step is 20, the parameter beta is 1.5 and the sample number is 10. For $L_2$ white-box attacks, the attack epsilon for all white-box attacks is 0.5. For PGD100, the attack step is 100, the attack step size is 1.0, and the attack starts with an random noise. For AutoAttack, we adopt the standard version of attack setting.

For adversarial training, we generate adversarial examples by PGD-3 attack, in which the attack epsilon is $\epsilon=4/255$, the attack step is $\mathrm{step}=3$ and the attack step size is calculated by $\mathrm{size}=2*\epsilon/\mathrm{step}$. The basic training setting for Transformers is based on the basic setting in Swin Transformers \citep{liu2021swin}. We adopt AdamW~\citep{loshchilov2017decoupled} optimizer with momentum 0.9 and weight decay 0.0. The learning scheduler is set with 300 epochs (without pre-trained models), cosine scheduler, learning rate 5e-4, 5 warm up epochs and decay rate 0.1. The input images are first center cropped with crop percentage $87.5\%$,  resized to image size 224 and normalized with mean $(0.485, 0.456, 0.406)$ and std $(0.229, 0.224, 0.225)$. Then the normalized images are augmented by color-jittering 0.4, RandAugment~\citep{cubuk2020randaugment} with parameter rand-m9-mstd0.5-inc1, Random Erasing with probability 0.25 and Mixup~\citep{zhang2017mixup}. We use soft label in label smoothing~\citep{szegedy2016rethinking} with parameter 0.1. No dropout used in adversarial training. The parameter for EMA~\citep{bolme2009average} is 0.9998. The training setting for ConvNext~\citep{liu2022convnet} is the same as that for Transformers. However, the basic training setting for ResNet~\citep{he2015deep} is different from that for Transformers, because some of the training strategies are not suitable for ResNet. We adopt SGD with momentum 0.9 and weight decay 0.0. The learning scheduler is set as 95 epochs, step scheduler, learning rate 0.2, 5 warm up epochs, decay rate 0.1 and 30 decay epochs. The rest settings are the same as these in Transformers. The batch size for adversarial training is 512.

\section{Results on ImageNet-C}\label{app:b}
\begin{figure*}[t]
  \centering
  \includegraphics[width=0.98\linewidth]{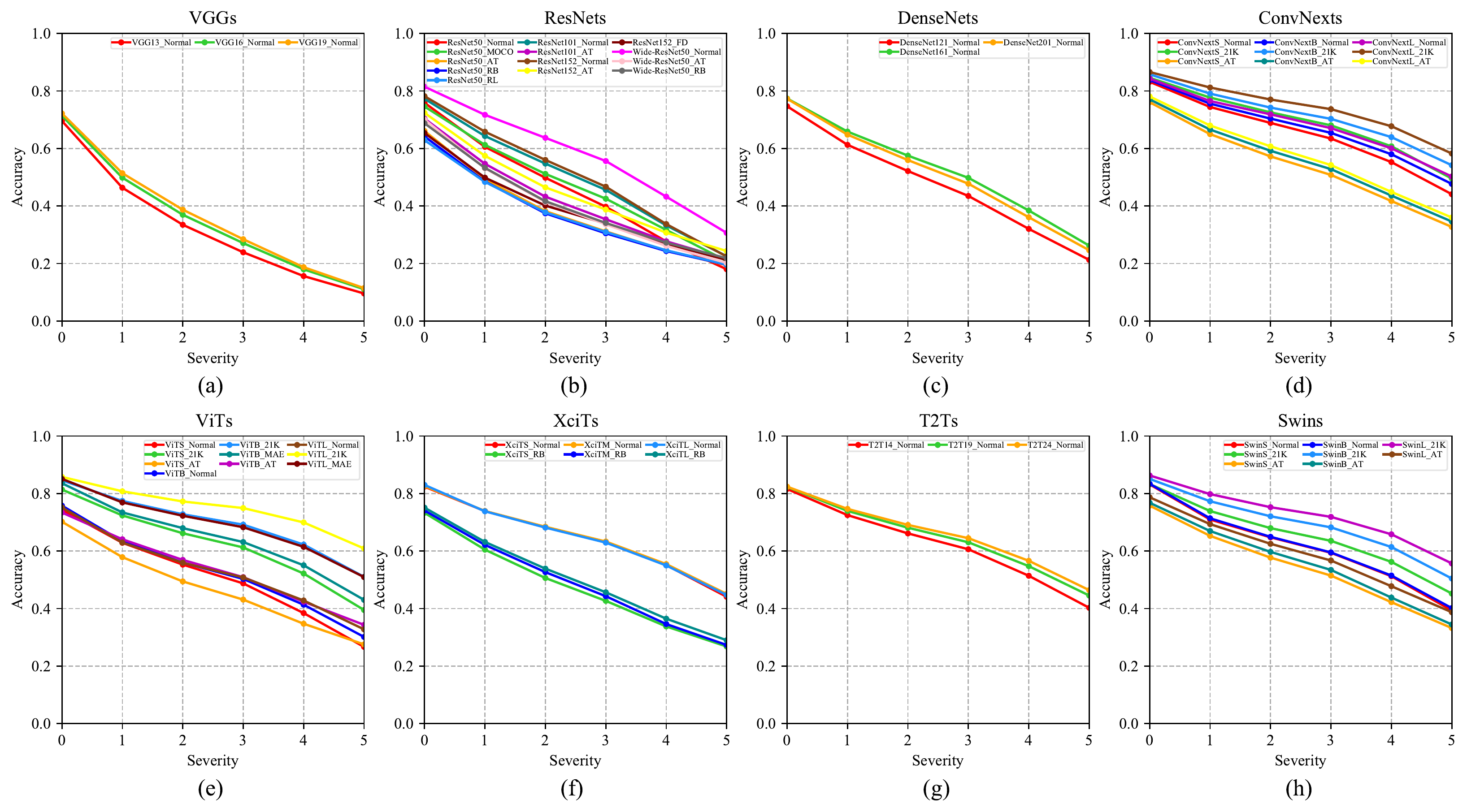}
  \caption{Robust curves on ImageNet-C. (a) Robust curves of VGGs (b) Robust curves of ResNets. (c) Robust curves of DenseNets. (d) Robust curves of ConvNexts. (e) Robust curves of ViTs. (f) Robust curves of XciTs. (g) Robust curves of T2Ts. (h) Robust curves of Swins. OOD robustness under different severity is generally positive correlated to clean accuracy, except for adversarially-trained models, which are more resistant to severe corruptions.}
  \label{fig:inc-app}
\end{figure*}
We also provide detailed robust curves of all the basic models in Tab.~\textcolor{blue}{2}. In this section, the robust curves are organized by model architectures. Detailed results of these robust curves are shown in Fig.~\ref{fig:inc-app}.

First, it is shown that 21K pre-training and SSL greatly improve natural robustness. Pre-training with 21K dataset prevent models from overfitting into certain distribution. SSL help model learn representative features, which benefit natural robustness. Second, it can be seen that the natural robustness will be improved with larger model size. Large model size means large model capacity, which benefits natural robustness. Third, the variation trends of normally-trained, 21K pre-trained and self-supervised pre-trained models are similar. However, the robust curves of adversarially-trained models become flat on large severity. It means adversarial training, in some degree, makes models resistant to more severe natural noise. This is owed to the fact that corruption noise with large severity damages the texture information. Adversarial training forces models to focus on shape-biased features, which can overcome the disturbance of severe noise.

\section{Result for different models}\label{app:c}
\begin{figure*}[t]
  \centering
  \includegraphics[width=0.98\linewidth]{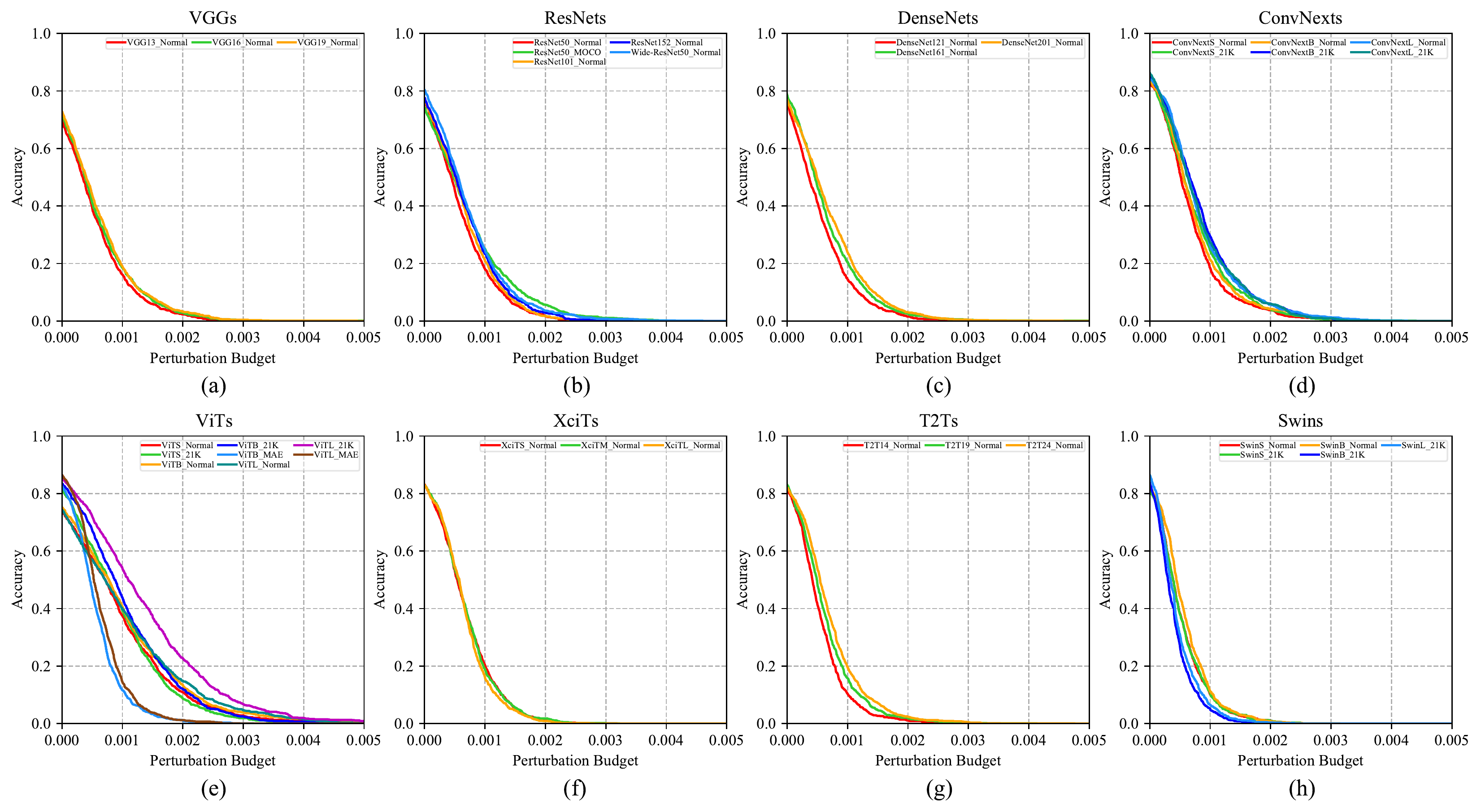}
  \caption{Robust curves of naturally-trained models (including normally-trained, 21K pre-trained and self-supervised pre-trained models) with AutoAttack. (a) Robust curves of VGGs (b) Robust curves of ResNets. (c) Robust curves of DenseNets. (d) Robust curves of ConvNexts. (e) Robust curves of ViTs. (f) Robust curves of XciTs. (g) Robust curves of T2Ts. (h) Robust curves of Swins. ViTL pre-trained with 21K dataset is the most robust model.}
  \label{fig:rb-normal-app}
\end{figure*}

In this section, we provide detailed robust curves of all the basic models. The models will be divided to two groups: naturally-trained models (including normally-trained, 21K pre-trained and self-supervised pre-trained models) and adversarially-trained ones. The robust curves of naturally-trained models are shown in Fig.~\ref{fig:rb-normal-app}.

\begin{figure*}[t]
  \centering
  \includegraphics[width=0.98\linewidth]{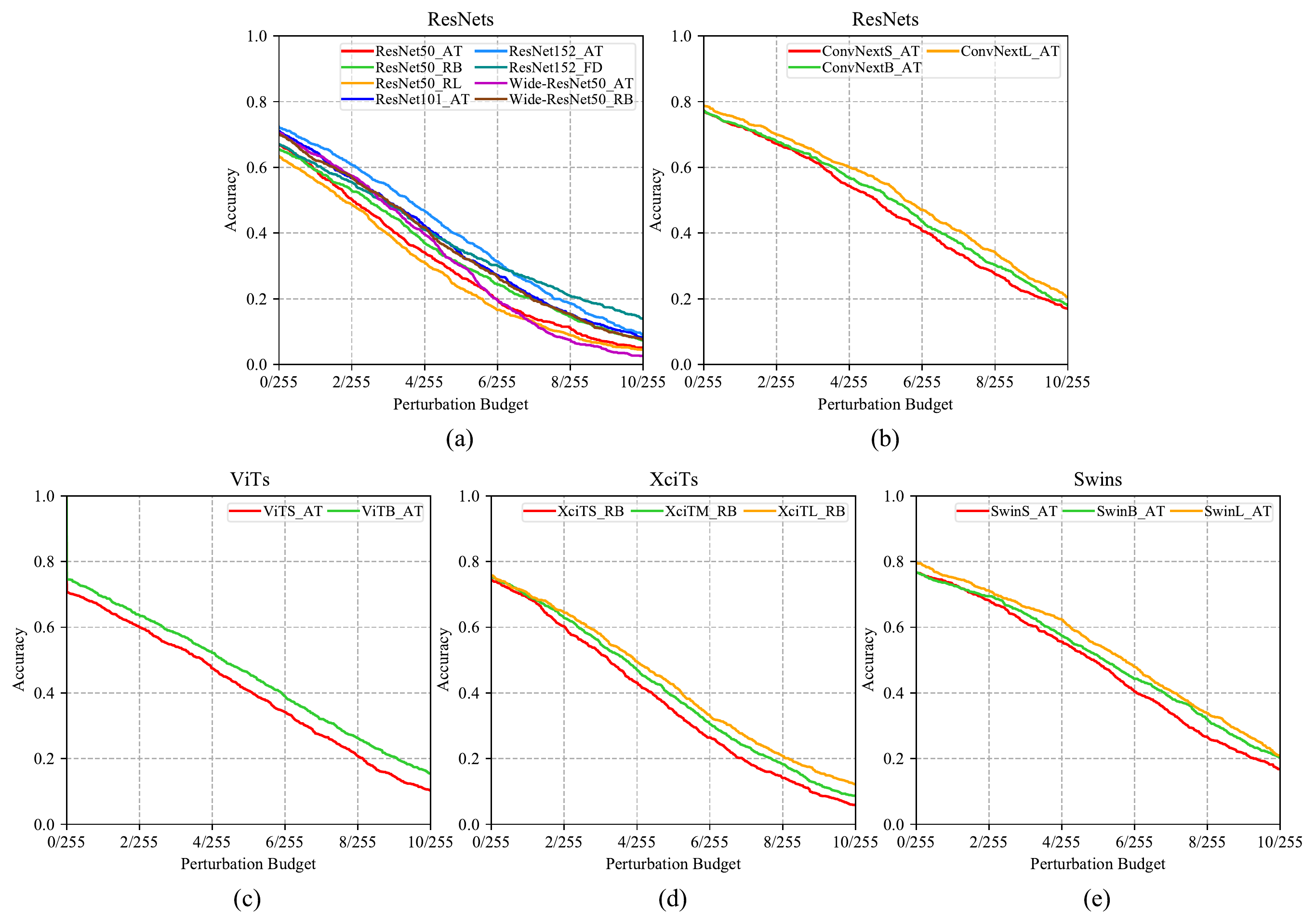}
  \caption{Robust curves of adversarially-trained models with AutoAttack. (a) Robust curves of ResNets (b) Robust curves of ConvNexts. (c) Robust curves of ViTs. (d) Robust curves of XciTs. (e) Robust curves of Swins.}
  \label{fig:rb-adv-app}
\end{figure*}

First, it is shown that 21K pre-training improves adversarial robustness on ConvNexts and ViTs~\citep{dosovitskiy2021image}. However, the robustness on Swins will decrease with 21K pre-training. Second, MAE~\citep{he2022masked} improves adversarial robustness on ViTs, while MOCOv3~\citep{chen2021empirical} benefits adversarial robustness. Finally, model size is less correlated to adversarial robustness. It is owed to the fact that normally-trained models have poor adversarial robustness. Therefore, the robust curves of normally-trained models are so close to each other.

The robust curves of adversarially-trained models are shown in Fig.~\ref{fig:rb-adv-app}. It is shown that the robust curves of the models with the same architecture are similar. First, ConvNext Large performs best in CNNs, and Swin Transformer Large performs best in Transformers. They all benefit from modern architecture designs, such as patchified input images, enlarged kernel size, and reduced activation and normalization layers~\citep{wang2022can}. Second, there exists trade-off between different perturbation budgets. The ResNet152\_FD is trained with epsilon 8/255, while the other models are trained with epsilon 4/255. This leads to the worse performance of ResNet152\_FD with perturbation budget smaller than 4/255. Third, larger model size generally means better adversarial robustness. With adversarial training, the robust curves of the models with similar architectures become parallel to each other. In this way, models' capacity also have effect on the adversarial robustness. Besides, the depth design of large models, such as SwinL (with depth 2-2-18-2) and ConvNextL (with depth 3-3-27-3) is consistent with the conclusion in~\citep{https://doi.org/10.48550/arxiv.2110.03825} that reducing capacity at the last stage benefits adversarial robustness.

\section{White-box attacks}\label{app:d}
Apart from $L_\infty$ norm, we also provide white-box evaluation with adversarial examples with $L_2$ norm. We only evaluate models with FGSM, PGD100 and AutoAttack. The epsilon for each attack method is set to 0.5. The detailed results are shown in Tab.~\ref{tab:appl2}. The conclusions are similar to these from $L_\infty$ attacks. ConvNextL is the most robust model in CNNs, and SwinL is the most robust model in Transformers.

\begin{table*}[h]\small
\begin{center}
\begin{minipage}{0.8\textwidth}
\caption{White-box robustness with different architectures. All the results are obtained by adversarial attacks with epsilon 0.5 and $L_2$ norm. ConvNextL is the most robust model in CNNs and SwinL is the most robust model in Transformers.}
\fontsize{7}{6}\selectfont 
\label{tab:appl2}
\begin{tabular*}{\textwidth}{@{\extracolsep{\fill}}llcccc@{\extracolsep{\fill}}}
\toprule%
Model & Source & Clean Acc &FGSM& PGD100 & AutoAttack \\
\midrule
\multirow{3}*{ResNet50} & AT & 67.0 &59.7& 58.7 & 56.0 \\
  & RB & 65.5 &57.2& 55.4 & 53.3 \\
 & RL & 63.4 &56.9& 55.9 & 54.0 \\
 \cmidrule{2-6}
ResNet101 & AT & 71.0 &64.3& 63.1 & 61.2 \\
 \cmidrule{2-6}
\multirow{2}*{ResNet152} & AT & 72.4&66.0 & 64.2 & 62.0 \\
 & FD & 62.0&61.2 & 60.2 & 57.2 \\
 \cmidrule{2-6}
\multirow{2}*{Wide-ResNet50} & AT & 70.5 &65.3& 64.1 & 61.9 \\
 & RB & 70.2 &62.0& 60.8 & 59.0 \\
 \midrule
 ConvNextS&AT&77.3&72.7&72.5&71.4\\
 \cmidrule{2-6}
  ConvNextB&AT&77.2&72.6&72.3&71.4\\
 \cmidrule{2-6}
  ConvNextL&AT&78.8&75.0&74.7&73.7\\
 \midrule
SwinS & AT & 75.8 &72.3&72.2 & 71.6 \\
 \cmidrule{2-6}
SwinB & AT & 76.6 &73.3& 73.1 & 72.6 \\
 \cmidrule{2-6}
SwinL & AT & 79.7 &75.4& 75.3 & 74.7 \\
\midrule
ViTS & AT & 70.7 &65.4 & 65.1 & 63.6 \\
 \cmidrule{2-6}
ViTB & AT & 74.7 &70.3 & 70.2 & 69.5 \\
\midrule
XciTS & RB & 74.5&70.7 & 70.5 & 69.8 \\
 \cmidrule{2-6}
XciTM & RB & 75.3&71.8 & 71.6 & 70.9 \\
 \cmidrule{2-6}
XciTL & RB & 75.8&71.7 & 71.6 & 71.4 \\
\botrule
\end{tabular*}
\end{minipage}
\end{center}
\end{table*}

\section{Black-box attacks}\label{app:e}
\begin{figure*}[t]
  \centering
  \includegraphics[width=\linewidth]{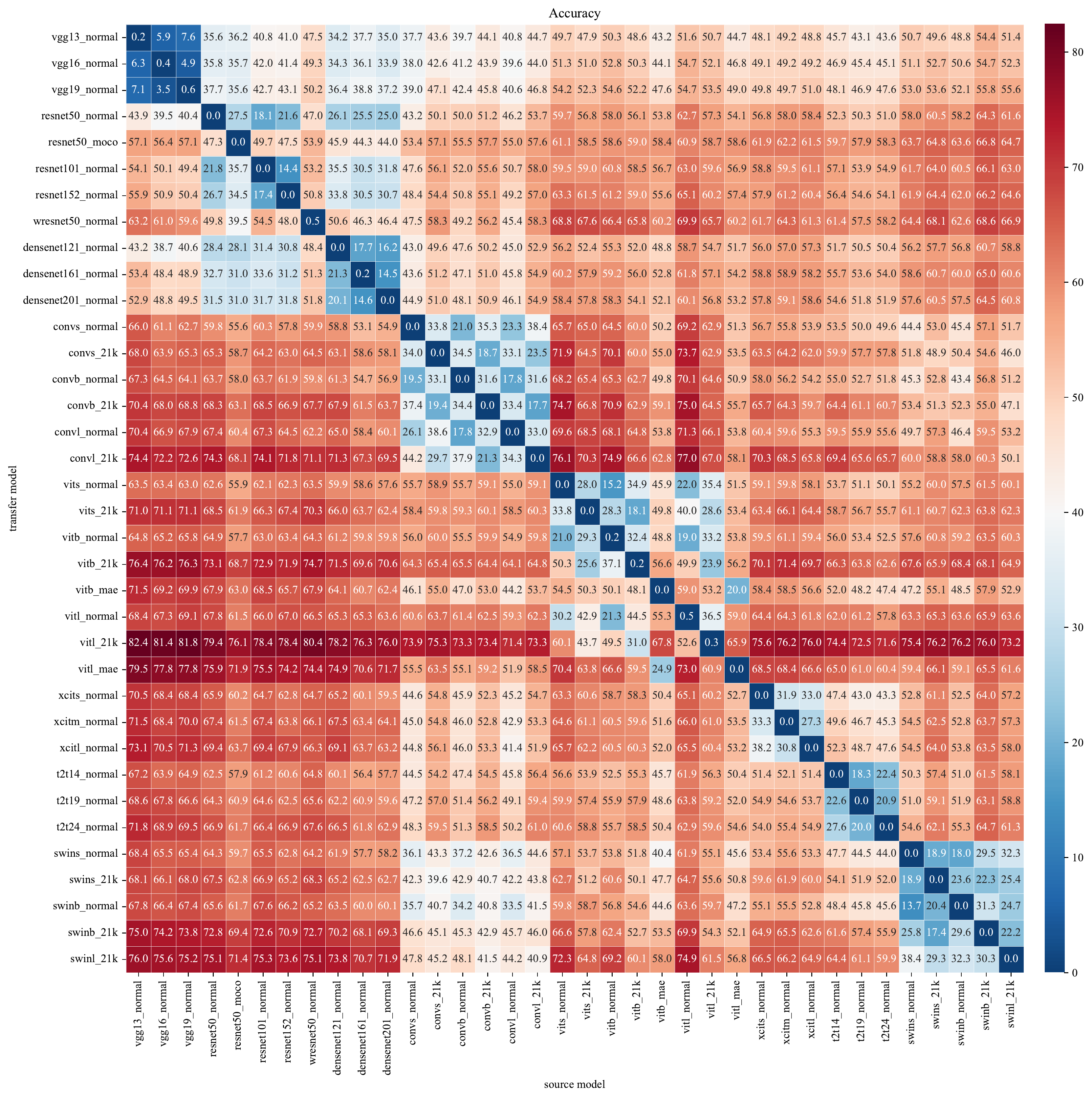}
  \caption{Black-box transferability of naturally-trained models, including normally-trained, 21K pre-trained and self-supervised pre-trained models, with black-box attack MIM.}
  \label{fig:black-natural-mim-app}
\end{figure*}

\begin{figure*}[t]
  \centering
  \includegraphics[width=\linewidth]{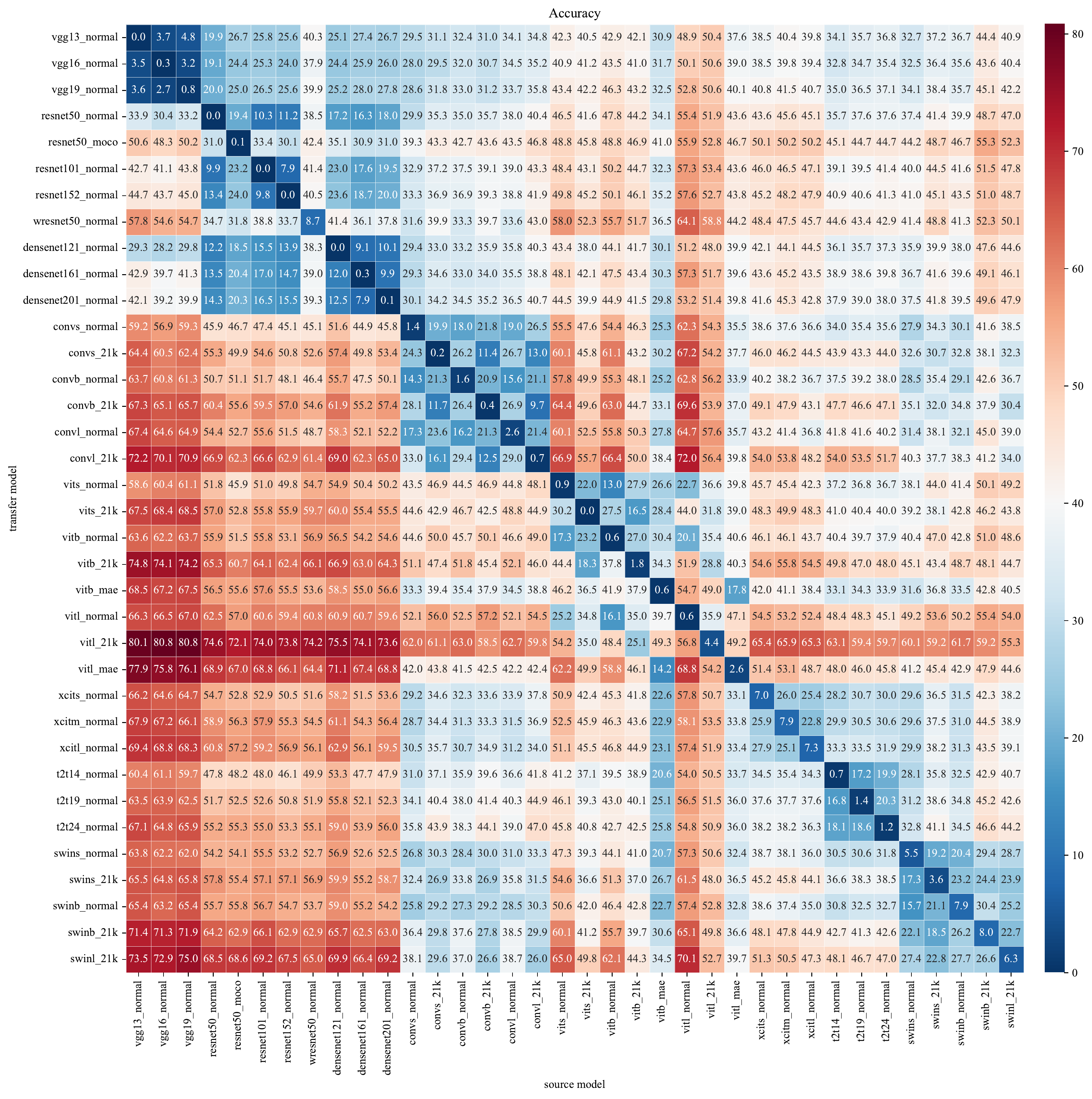}
  \caption{Black-box transferability of naturally-trained models, including normally-trained, 21K pre-trained and self-supervised pre-trained models, with black-box attack DIM.}
  \label{fig:black-natural-dim-app}
\end{figure*}

\begin{figure*}[t]
  \centering
  \includegraphics[width=\linewidth]{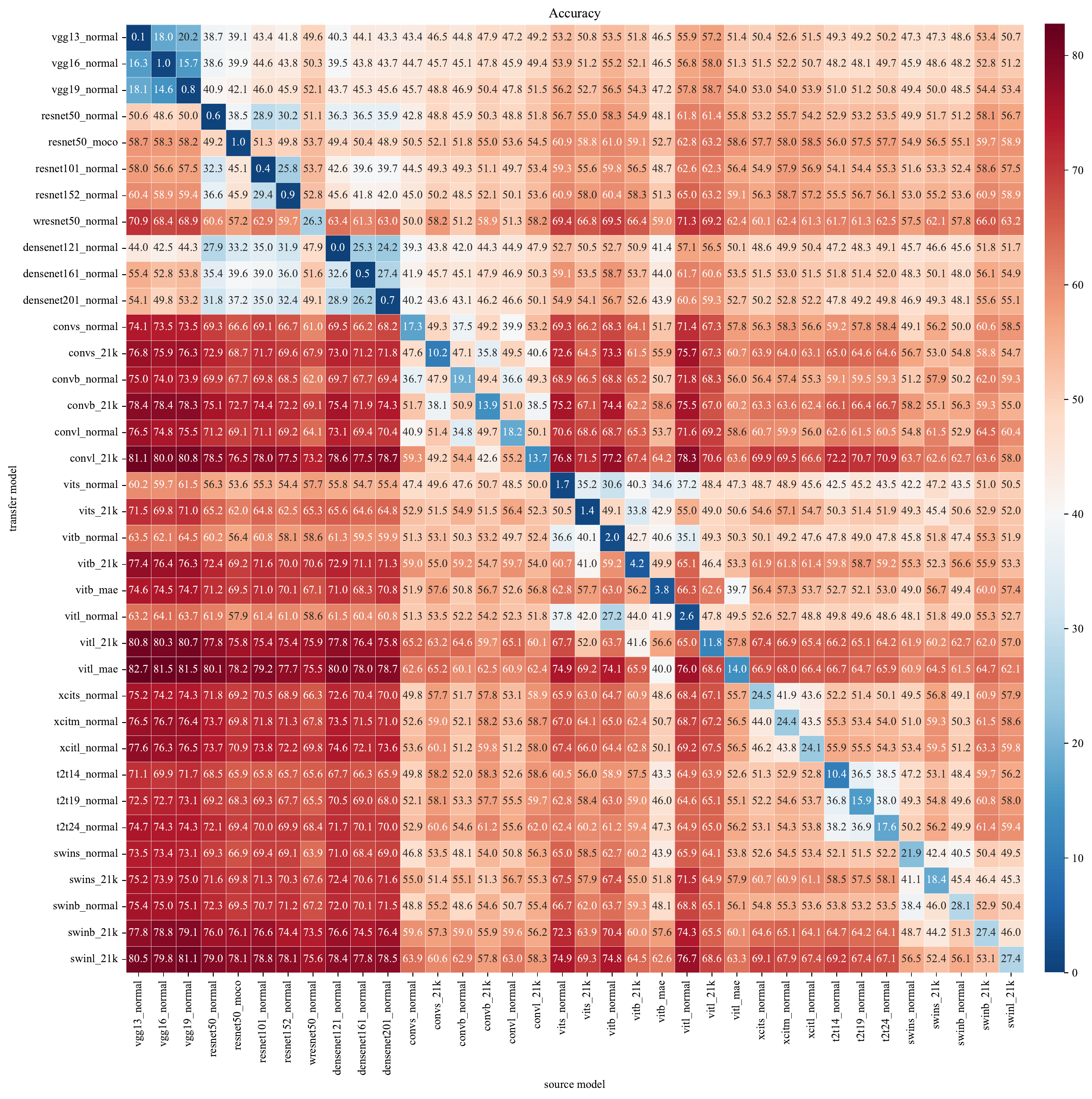}
  \caption{Black-box transferability of naturally-trained models, including normally-trained, 21K pre-trained and self-supervised pre-trained models, with black-box attack TIM.}
  \label{fig:black-natural-tim-app}
\end{figure*}

\begin{figure*}[t]
  \centering
  \includegraphics[width=\linewidth]{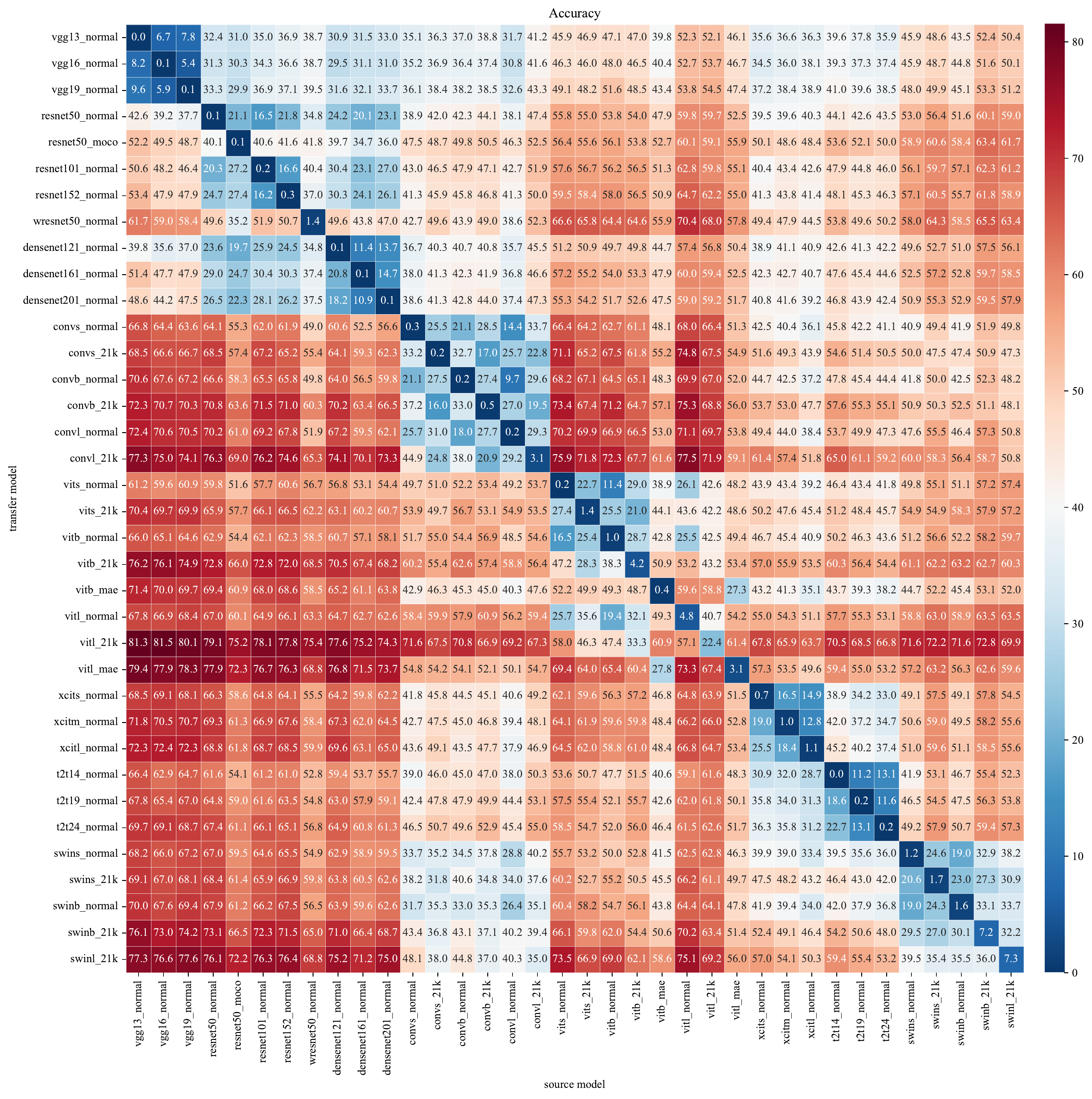}
  \caption{Black-box transferability of naturally-trained models, including normally-trained, 21K pre-trained and self-supervised pre-trained models, with black-box attack SI-NI-FGSM.}
  \label{fig:black-natural-si-app}
\end{figure*}

\begin{figure}[t]
  \centering
  \includegraphics[width=\linewidth]{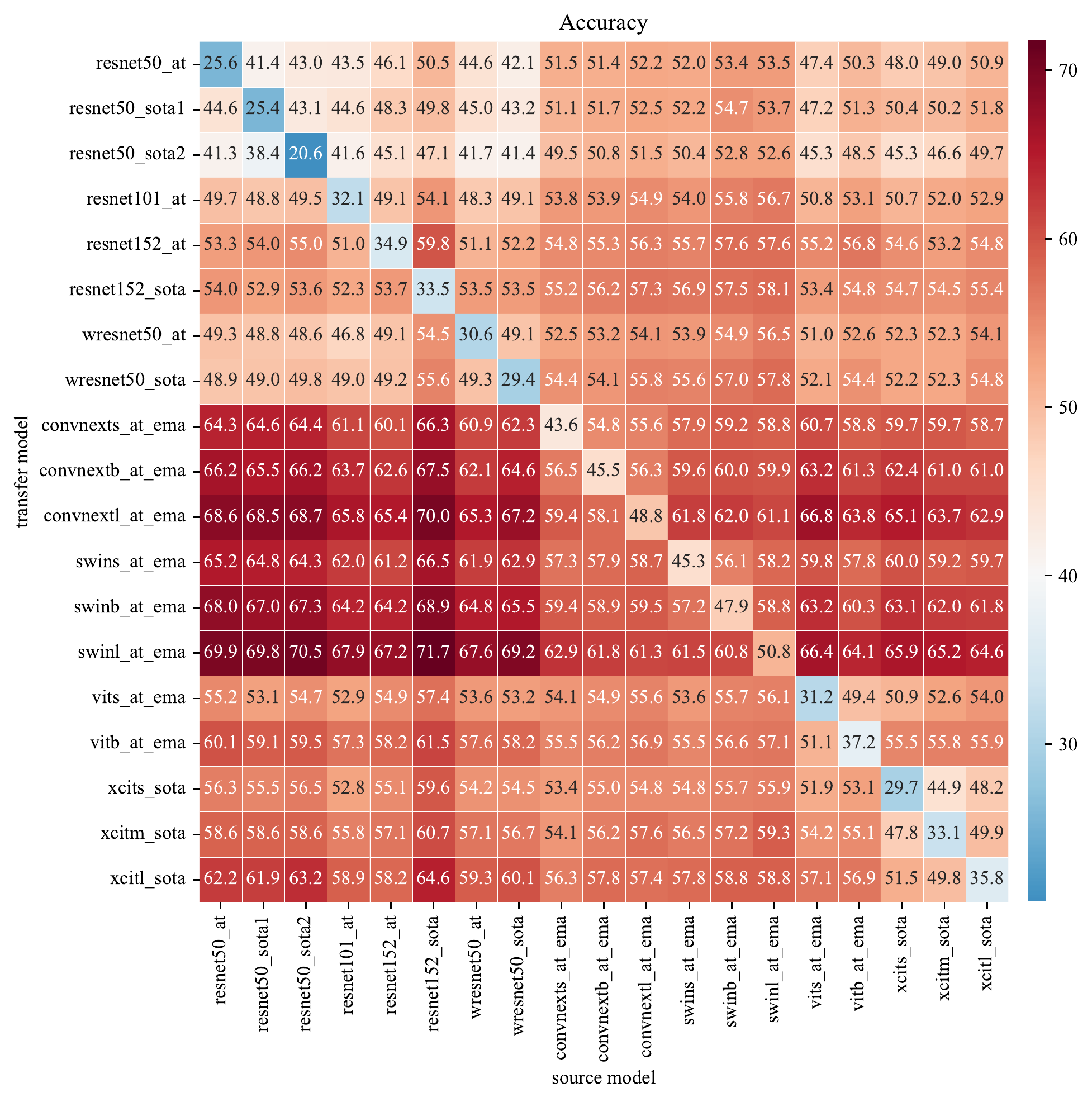}
  \caption{Black-box transferability of adversarially trained models with black-box attack MIM.}
  \label{fig:transfer-adv-mim-app}
\end{figure}

\begin{figure}[t]
  \centering
  \includegraphics[width=\linewidth]{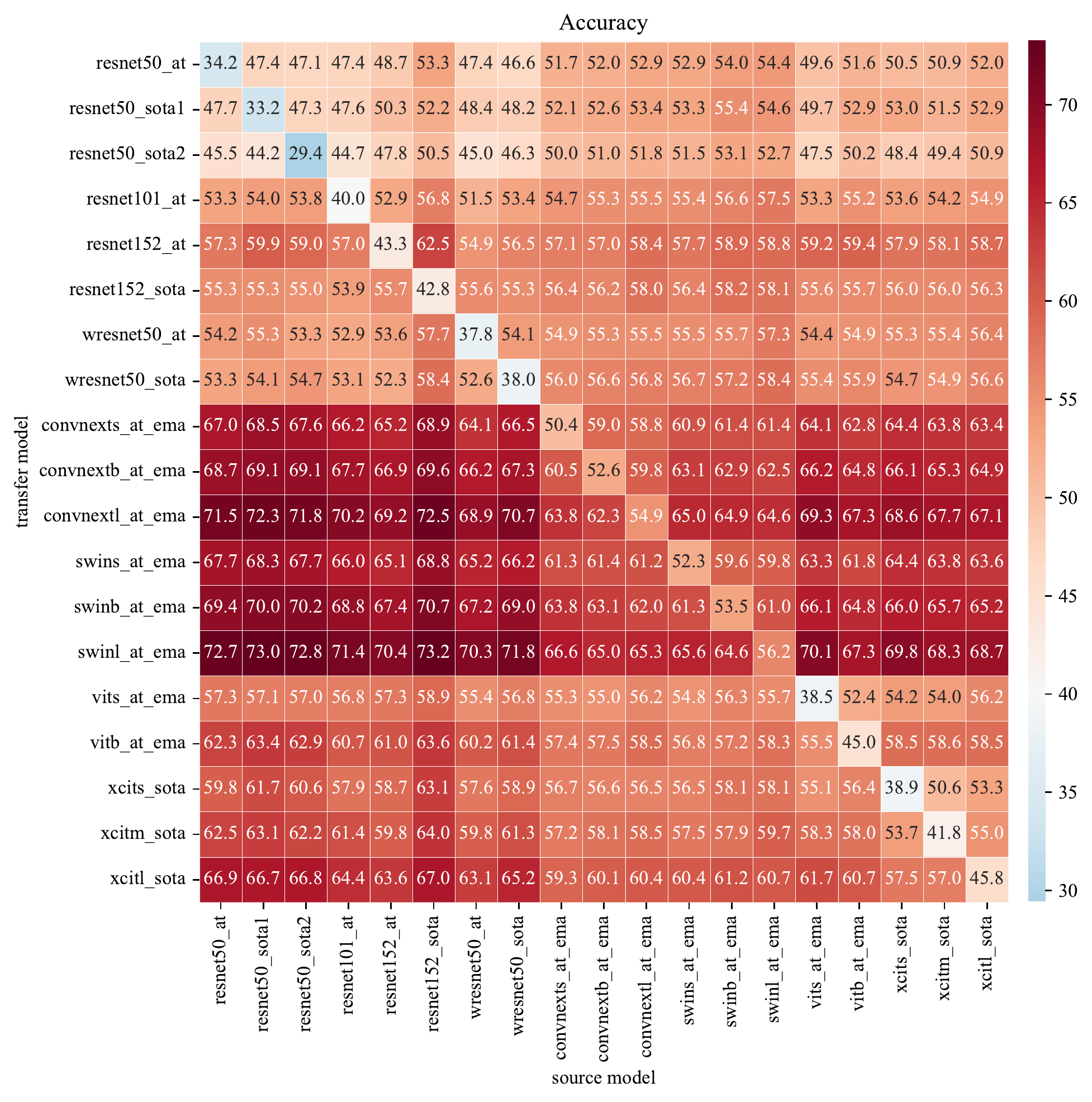}
  \caption{Black-box transferability of adversarially trained models with black-box attack DIM.}
  \label{fig:transfer-adv-dim-app}
\end{figure}

\begin{figure}[t]
  \centering
  \includegraphics[width=\linewidth]{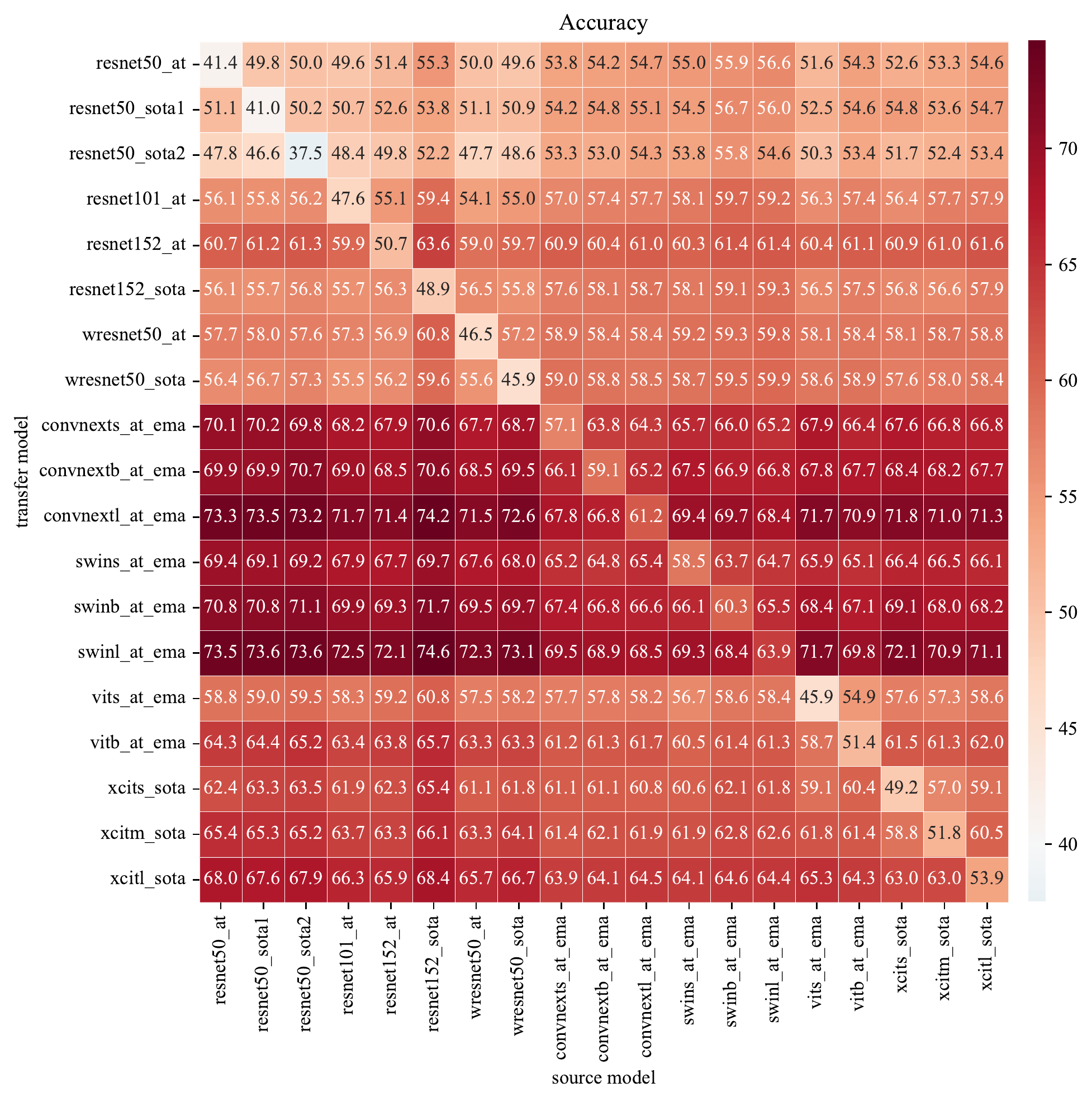}
  \caption{Black-box transferability of adversarially trained models with black-box attack TIM.}
  \label{fig:transfer-adv-tim-app}
\end{figure}

\begin{figure}[t]
  \centering
  \includegraphics[width=\linewidth]{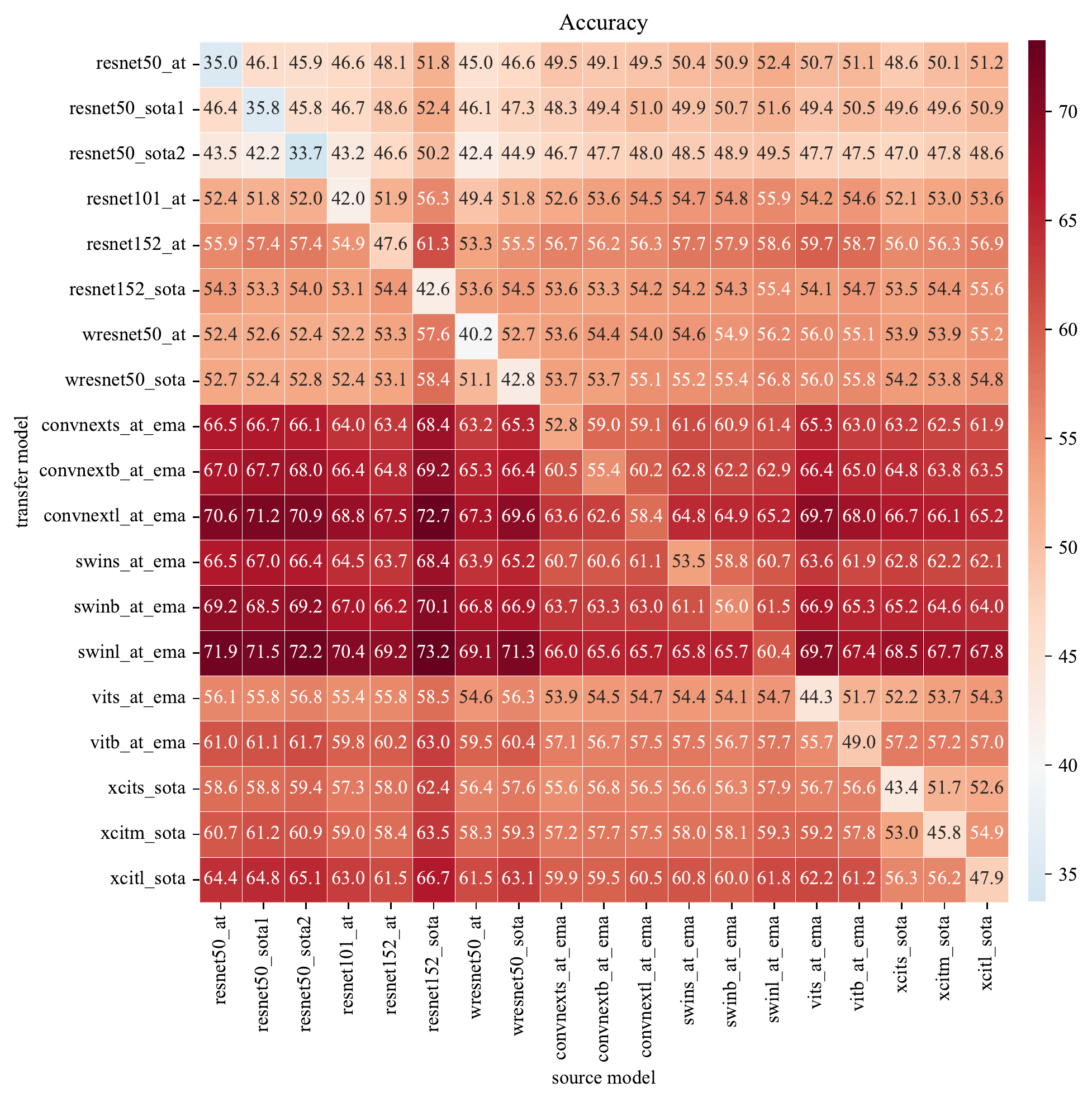}
  \caption{Black-box transferability of adversarially trained models with black-box attack SI-NI-FGSM.}
  \label{fig:transfer-adv-si-app}
\end{figure}

We also evaluate the black-box robustness of different models with more adversarial attacks, including MIM, DIM, TIM, and SI-NI-FGSM. The detailed results of naturally-trained models with adversarial attacks, including MIM, DIM, TIM, and SI-NI-FGSM, are shown in Fig.~\ref{fig:black-natural-mim-app}-Fig.~\ref{fig:black-natural-si-app}. The detailed results of adversarially-trained models with MIM, DIM, TIM, and SI-NI-FGSM are shown in Fig.~\ref{fig:transfer-adv-mim-app}-Fig.~\ref{fig:transfer-adv-si-app}.

First, ViT models are generally more robust than other models under the black-box setting. Second, ConvNext models act as the best surrogate models. Third, naturally-trained models are generally less robust on DIM attacks. While adversarially-trained models are generally less robust on MIM attacks. This is because, transferability is not the decisive factor to attack an adversarially-trained models, while the attack ability to surrogate models is also important. Fourth, models in one architecture are relatively more robust to adversarial examples generated by the surrogate models in another architecture.






\end{appendices}



\renewcommand*{\bibfont}{\small} 
\bibliography{sn-bibliography}